\documentclass[10pt,twocolumn,letterpaper]{article}

\usepackage[pagenumbers]{cvpr} 

\usepackage{graphicx}
\usepackage{array}
\usepackage{multirow}
\usepackage{amsmath}
\usepackage{amssymb}
\usepackage{textcomp}  
\usepackage{gensymb}
\usepackage{booktabs}

\usepackage{xcolor}  
\usepackage{siunitx}  
\usepackage{appendix}

\usepackage{microtype}

\usepackage{pifont}
\newcommand{\cmark}{\ding{51}}%
\definecolor{citecolor}{HTML}{65C3A6}
\usepackage[pagebackref,breaklinks,colorlinks,citecolor=citecolor]{hyperref}

\usepackage[capitalize]{cleveref}
\crefname{section}{Sec.}{Secs.}
\Crefname{section}{Section}{Sections}
\Crefname{table}{Table}{Tables}
\crefname{table}{Tab.}{Tabs.}

\newif\ifsubmit
\submittrue

\newcommand{\cutsectionup}{\vspace*{-2pt}} 
\newcommand{\cutsectiondown}{\vspace*{-2pt}}
\newcommand{\cutsubsectionup}{\vspace*{-1pt}}
\newcommand{\cutsubsectiondown}{\vspace*{-1pt}}

\newcommand{\cutparagraphup}{\vspace*{-5pt}}

\newcommand{\cuthalfcaptionup}{\vspace*{-5pt}}

\newcommand{\cutabstractup}{\vspace*{-10pt}}

\begin{document}

\definecolor{yellow}{rgb}{1,1, 0.6}
\definecolor{lightyellow}{rgb}{1,1, 0.8}
\definecolor{orange}{rgb}{1, 0.8, 0.6}
\definecolor{red}{rgb}{1, 0.6, 0.6}

\definecolor{wincolor}{rgb}{0.85, 0.0, 0.0}

\definecolor{darkyellow}{rgb}{0.8, 0.8, 0.5}
\definecolor{darkred}{rgb}{0.7, 0.3, 0.3}
\definecolor{darkgreen}{rgb}{0.3, 0.7, 0.3}
\definecolor{blue}{rgb}{0, 0, 1.0}
\definecolor{green}{rgb}{0, 1.0, 0}
\definecolor{pink}{rgb}{1, 0.4, 0.7}

\newcommand{\matt}[1]{{\color{blue} Matt: #1}}

\newcommand{\modeltheta}{\mathrm{\Theta}}
\newcommand{\absrp}{\sigma}

\newcommand{\numsamples}{N}
\newcommand{\numsamplescoarse}{N_c}
\newcommand{\numsamplesfine}{N_f}

\newcommand{\posxy}{xy}
\newcommand{\posxyz}{xyz}
\newcommand{\angletheta}{\theta}
\newcommand{\anglephi}{\phi}
\newcommand{\posall}{\posxyz\angletheta\anglephi}

%
%

\title{Block-NeRF: Scalable Large Scene Neural View Synthesis}

\author{
Matthew Tancik$^{1*}$
\qquad
Vincent Casser$^2$
\qquad
Xinchen Yan$^2$
\qquad
Sabeek Pradhan$^2$\\
Ben Mildenhall$^3$
\qquad
Pratul P. Srinivasan$^3$
\qquad
Jonathan T. Barron$^3$
\qquad
Henrik Kretzschmar$^2$\\[9pt]
$^1$UC Berkeley \qquad $^2$Waymo \qquad $^3$Google Research
}

\twocolumn[{%
\renewcommand\twocolumn[1][]{#1}%
\maketitle
\vspace{-18pt}
\begin{center}
\captionsetup{type=figure}
\cuthalfcaptionup
    \includegraphics[width=\linewidth]{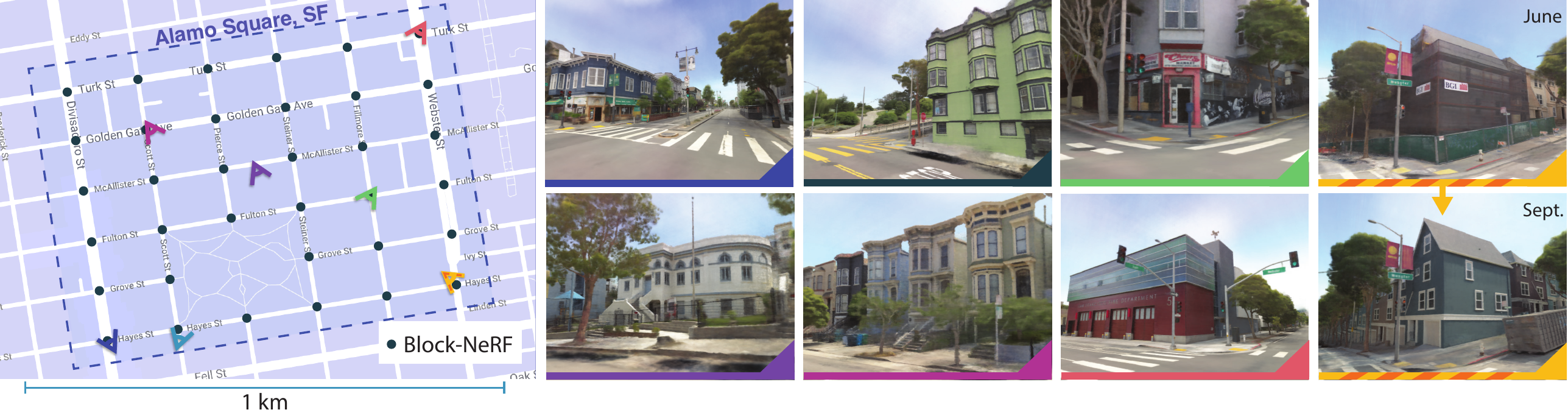}
    \vspace{-20pt}
\captionof{figure}{\textbf{Block-NeRF} is a method that enables large-scale scene reconstruction by representing the environment using multiple compact NeRFs that each fit into memory. At inference time, Block-NeRF seamlessly combines renderings of the relevant NeRFs for the given area. In this example, we reconstruct the Alamo Square neighborhood in San Francisco using data collected over 3 months. Block-NeRF can update individual blocks of the environment without retraining on the entire scene, as demonstrated by the construction on the right. Video results can be found on the project website \href{http://waymo.com/research/block-nerf}{waymo.com/research/block-nerf}.}
\cuthalfcaptionup
\label{fig:teaser}
\end{center}%
}]

\let\thefootnote\relax\footnote{*Work done as an intern at Waymo.}

\begin{abstract}
We present Block-NeRF, a variant of Neural Radiance Fields that can represent large-scale environments.
Specifically, we demonstrate that when scaling NeRF to render city-scale scenes spanning multiple blocks, it is vital to decompose the scene into individually trained NeRFs. This decomposition decouples rendering time from scene size, enables rendering to scale to arbitrarily large environments, and allows per-block updates of the environment. We adopt several architectural changes to make NeRF robust to data captured over months under different environmental conditions.
We add appearance embeddings, learned pose refinement, and controllable exposure to each individual NeRF, and introduce a procedure for aligning appearance between adjacent NeRFs so that they can be seamlessly combined. We build a grid of Block-NeRFs from 2.8 million images to create the largest neural scene representation to date, capable of rendering an entire neighborhood of San Francisco.

\end{abstract}

\cutabstractup

\cutsectionup
\section{Introduction}
Recent advancements in neural rendering such as Neural Radiance Fields~\cite{mildenhall2020nerf} have enabled photo-realistic reconstruction and novel view synthesis given a set of posed camera images~\cite{barron2021mip,park2020deformable,martin2021nerf}. 
Earlier works tended to focus on small-scale and object-centric reconstruction. 
Though some methods now address scenes the size of a single room or building, these are generally still limited and do not na\"ively scale up to \textit{city-scale} environments. Applying these methods to large environments typically leads to significant artifacts and low visual fidelity due to limited model capacity.

Reconstructing large-scale environments enables several important use-cases in domains such as autonomous driving~\cite{ost2021neural,li2019aads,yang2020surfelgan} and aerial surveying~\cite{du2018unmanned,liu2021infinite}.
One example is mapping, where a high-fidelity map of the entire operating domain is created to act as a powerful prior for a variety of problems, including robot localization, navigation, and collision avoidance. 
Furthermore, large-scale scene reconstructions can be used for closed-loop robotic simulations~\cite{dosovitskiy2017carla}. Autonomous driving systems are commonly evaluated by re-simulating previously encountered scenarios; however, any deviation from the recorded encounter may change the vehicle's trajectory, requiring high-fidelity novel view renderings along the altered path. Beyond basic view synthesis, scene conditioned NeRFs are also capable of changing environmental lighting conditions such as camera exposure, weather, or time of day, which can be used to further augment simulation scenarios.

Reconstructing such large-scale environments introduces additional challenges, including the presence of transient objects (cars and pedestrians), limitations in model capacity, along with memory and compute constraints. Furthermore, training data for such large environments is highly unlikely to be collected in a single capture under consistent conditions. Rather, data for different parts of the environment may need to be sourced from different data collection efforts, introducing variance in both scene geometry (\eg, construction work and parked cars), as well as appearance (\eg, weather conditions and time of day).

We extend NeRF with appearance embeddings and learned pose refinement to address the environmental changes and pose errors in the collected data. We additionally add exposure conditioning to provide the ability to modify the exposure during inference. We refer to this modified model as a Block-NeRF. Scaling up the network capacity of Block-NeRF enables the ability to represent increasingly large scenes. However this approach comes with a number of limitations; rendering time scales with the size of the network, networks can no longer fit on a single compute device, and updating or expanding the environment requires retraining the entire network.

To address these challenges, we propose dividing up large environments into individually trained Block-NeRFs, which are then rendered and combined dynamically at inference time. Modeling these Block-NeRFs independently allows for maximum flexibility, scales up to arbitrarily large environments and provides the ability to update or introduce new regions in a piecewise manner without retraining the entire environment as demonstrated in Figure~\ref{fig:teaser}. To compute a target view, only a subset of the Block-NeRFs are rendered and then composited based on their geographic location compared to the camera. To allow for more seamless compositing, we propose an appearance matching technique which brings different Block-NeRFs into visual alignment by optimizing their appearance embeddings.

\cutsectionup
\section{Related Work}
\cutsubsectiondown

\begin{figure}[t]
    \centering
    \includegraphics[width=.8\linewidth]{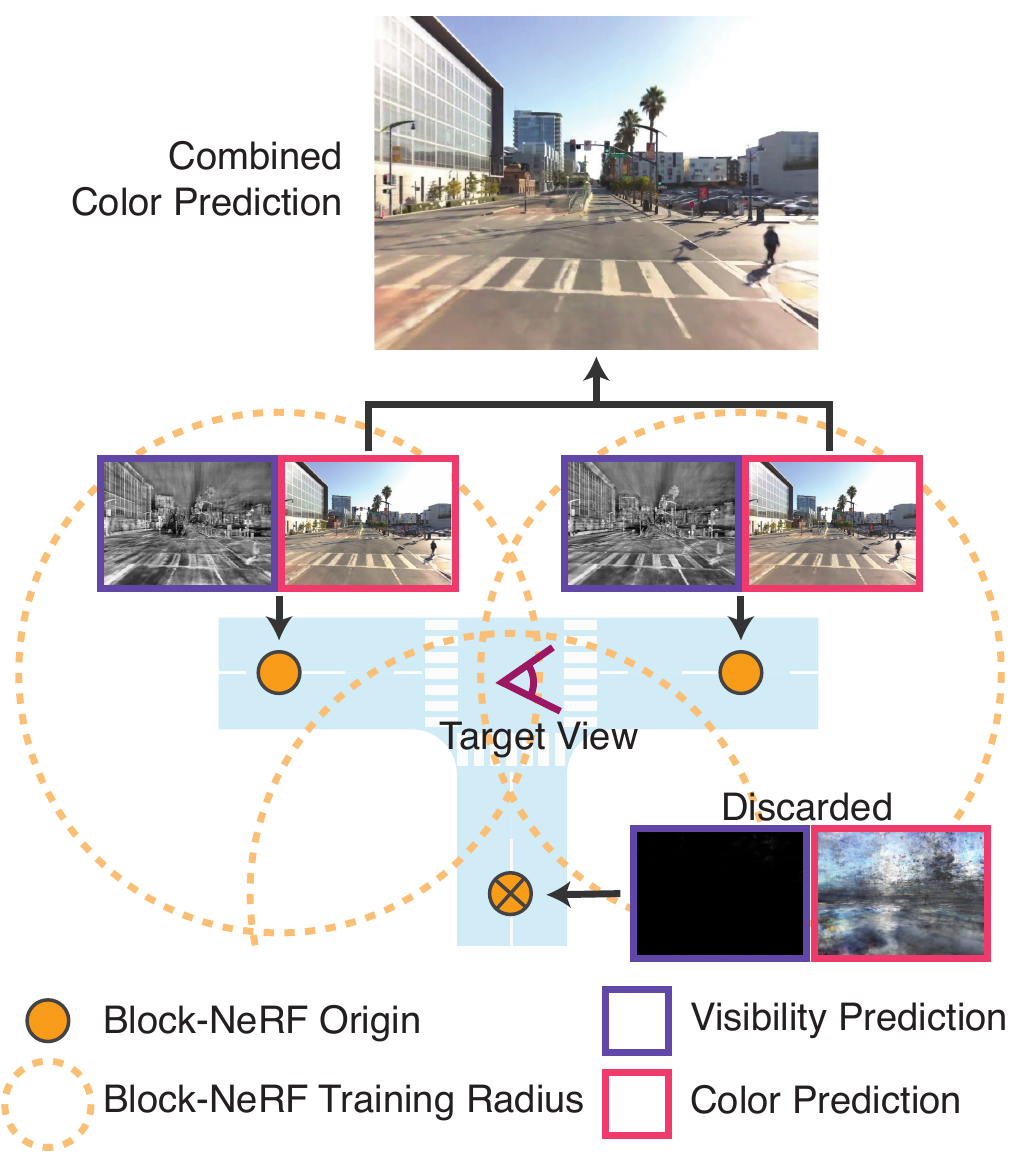}
    \cuthalfcaptionup
    \caption{The scene is split into multiple Block-NeRFs that are each trained on data within some radius (dotted orange line) of a specific Block-NeRF origin coordinate (orange dot). To render a target view in the scene, the visibility maps are computed for all of the NeRFs within a given radius. Block-NeRFs with low visibility are discarded (bottom Block-NeRF) and the color output is rendered for the remaining blocks. The renderings are then merged based on each block origin's distance to the target view. }
    \label{fig:merge}
    \cuthalfcaptionup
    \cuthalfcaptionup
\end{figure}

\cutsubsectionup
\subsection{Large Scale 3D Reconstruction}
\cutsubsectiondown

Researchers have been developing and refining techniques for 3D reconstruction from large image collections for decades~\cite{fruh2004automated,snavely2006phototourism,pollefeys2008detailed,li2008modeling,agarwal2011building,zhu2018very}, and much current work relies on mature and robust software implementations such as COLMAP to perform this task~\cite{schonberger2016structure}. Nearly all of these reconstruction methods share a common pipeline: extract 2D image features (such as SIFT~\cite{lowe2004sift}), match these features across different images, and jointly optimize a set of 3D points and camera poses to be consistent with these matches (the well-explored problem of bundle adjustment~\cite{hartleyziss2004,triggs1999bundle}). Extending this pipeline to city-scale data is largely a matter of implementing highly robust and parallelized versions of these algorithms, as explored in work such as Photo Tourism~\cite{snavely2006phototourism} and Building Rome in a Day~\cite{agarwal2011building}. Core graphics research has also explored breaking up scenes for fast high quality rendering~\cite{losasso2004geometry}.

These approaches typically output a camera pose for each input image and a sparse 3D point cloud. To get a complete 3D scene model, these outputs must be further processed by a dense multi-view stereo algorithm (\eg, PMVS~\cite{furukawa2010pmvs}) to produce a dense point cloud or triangle mesh. This process presents its own scaling difficulties~\cite{furukawa2010towards}. The resulting 3D models often contain artifacts or holes in areas with limited texture or specular reflections as they are challenging to triangulate across images. As such, they frequently require further postprocessing to create models that can be used to render convincing imagery~\cite{shan2013turing}. However, this task is mainly the domain of novel view synthesis, and 3D reconstruction techniques primarily focus on geometric accuracy.

In contrast, our approach does not rely on large-scale SfM to produce camera poses, instead performing odometry using various sensors on the vehicle as the images are collected~\cite{thrun2002probabilistic}.

\cutsubsectionup
\subsection{Novel View Synthesis}
\cutsubsectiondown
Given a set of input images of a given scene and their camera poses, novel view synthesis seeks to render observed scene content from previously unobserved viewpoints, allowing a user to navigate through a recreated environment with high visual fidelity. 
 
\cutparagraphup
\paragraph{Geometry-based Image Reprojection.}
Many approaches to view synthesis start by applying traditional 3D reconstruction techniques to build a point cloud or triangle mesh representing the scene. This geometric ``proxy'' is then used to reproject pixels from the input images into new camera views, where they are blended by heuristic~\cite{buehler2001unstructured} or learning-based methods~\cite{hedman2018deep,riegler2020fvs,riegler2021svs}. This approach has been scaled to long trajectories of first-person video~\cite{kopf2014hyperlapse}, panoramas collected along a city street~\cite{kopf2010street}, and single landmarks from the Photo Tourism dataset~\cite{meshry2019rerendering}.
Methods reliant on geometry proxies are limited by the quality of the initial 3D reconstruction, which hurts their performance in scenes with complex geometry or reflectance effects.

\cutparagraphup
\paragraph{Volumetric Scene Representations.}
Recent view synthesis work has focused on unifying reconstruction and rendering and learning this pipeline end-to-end, typically using a volumetric scene representation. 
Methods for rendering small baseline view interpolation often use feed-forward networks to learn a mapping directly from input images to an output volume~\cite{flynn2016deepstereo,zhou2018stereo}, while methods such as Neural Volumes~\cite{lombardi2019neuralvolumes} that target larger-baseline view synthesis run a global optimization over all input images to reconstruct every new scene, similar to traditional bundle adjustment. 

Neural Radiance Fields (NeRF)~\cite{mildenhall2020nerf} combines this single-scene optimization setting with a neural scene representation capable of representing complex scenes much more efficiently than a discrete 3D voxel grid; however, its rendering model scales very poorly to large-scale scenes in terms of compute.
Followup work has proposed making NeRF more efficient by partitioning space into smaller regions, each containing its own lightweight NeRF network~\cite{rebain2021derf,reiser2021kilonerf}.
Unlike our method, these network ensembles must be trained jointly, limiting their flexibility.
Another approach is to provide extra capacity in the form of a coarse 3D grid of latent codes~\cite{liu2020nsvf}. This approach has also been applied to compress detailed 3D shapes into neural signed distance functions~\cite{takikawa2021nglod} and to represent large scenes using occupancy networks~\cite{peng2020convolutional}.

We build our Block-NeRF implementation on top of mip-NeRF~\cite{barron2021mip}, which improves aliasing issues that hurt NeRF's performance in scenes where the input images observe the scene from many different distances. We incorporate techniques from NeRF in the Wild (NeRF-W)~\cite{martin2021nerf}, which adds a latent code per training image to handle inconsistent scene appearance when applying NeRF to landmarks from the Photo Tourism dataset.
NeRF-W creates a separate NeRF for each landmark from thousands of images, whereas our approach combines many NeRFs to reconstruct a coherent large environment from \emph{millions} of images. Our model also incorporates a learned camera pose refinement which has been explored in previous works~\cite{yariv2020multiview,su2021nerf,lin2021barf,wang2021nerf,yen2020inerf}.

Some NeRF-based methods use segmentation data to isolate and reconstruct static~\cite{yang2021learning} or moving objects (such as people or cars)~\cite{zhang2021editable,ost2021neural} across video sequences. As we focus primarily on reconstructing the environment itself, we choose to simply mask out dynamic objects during training.

 \begin{figure}
     \centering
     \includegraphics[width=\linewidth]{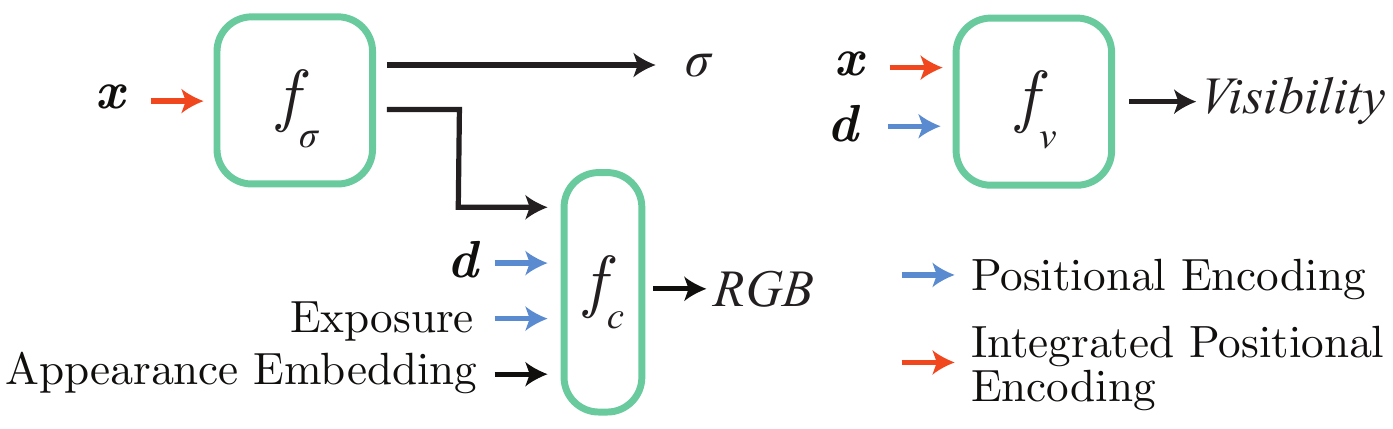}
     \caption{Our model is an extension of the model presented in mip-NeRF~\cite{barron2021mip}. The first MLP $f_\sigma$ predicts the density $\sigma$ for a position $\mathbf{x}$ in space. The network also outputs a feature vector that is concatenated with viewing direction $\mathbf{d}$, the exposure level, and an appearance embedding. These are fed into a second MLP $f_c$ that outputs the color for the point. We additionally train a visibility network $f_v$ to predict whether a point in space was visible in the training views, which is used for culling Block-NeRFs during inference.
     }
     \cuthalfcaptionup
     \label{fig:network}
 \end{figure}

\cutsubsectionup
\subsection{Urban Scene Camera Simulation}
\cutsubsectiondown
Camera simulation has become a popular data source for training and validating autonomous driving systems on interactive platforms~\cite{amini2020learning,kim2021drivegan}. Early works~\cite{gaidon2016virtual,richter2016playing,ros2016synthia,dosovitskiy2017carla} synthesized data from scripted scenarios and manually created 3D assets. These methods suffered from domain mismatch and limited scene-level diversity. Several recent works tackle the simulation-to-reality gaps by minimizing the distribution shifts in the simulation and rendering pipeline. Kar~\etal\cite{kar2019meta} and Devaranjan~\etal\cite{devaranjan2020meta} proposed to minimize the scene-level distribution shift from rendered outputs to real camera sensor data through a learned scenario generation framework. Richter~\etal\cite{richter2021enhancing} leveraged intermediate rendering buffers in the graphics pipeline to improve photorealism of synthetically generated camera images.

Towards the goal of building photo-realistic and scalable camera simulation, prior methods~\cite{li2019aads,yang2020surfelgan,chen2021geosim} leverage rich multi-sensor driving data collected during a single drive to reconstruct 3D scenes for object injection~\cite{chen2021geosim} and novel view synthesis~\cite{yang2020surfelgan} using modern machine learning techniques, including image GANs for 2D neural rendering.
Relying on a sophisticated surfel reconstruction pipeline, SurfelGAN~\cite{yang2020surfelgan} is still susceptible to errors in graphical reconstruction and can suffer from the limited range and vertical field-of-view of LiDAR scans.
In contrast to existing efforts, our work tackles the 3D rendering problem and is capable of modeling the real camera data captured from multiple drives under varying environmental conditions, such as weather and time of day, which is a prerequisite for reconstructing large-scale areas.
%
\cutsectionup
\section{Background}

We build upon NeRF~\cite{mildenhall2020nerf} and its extension mip-NeRF~\cite{barron2021mip}. Here, we summarize relevant parts of these methods. For details, please refer to the original papers.
\cutsubsectionup
\subsection{NeRF and mip-NeRF Preliminaries}
\cutsubsectiondown
Neural Radiance Fields (NeRF)~\cite{mildenhall2020nerf} is a coordinate-based neural scene representation that is optimized through a differentiable rendering loss to reproduce the appearance of a set of input images from known camera poses. After optimization, the NeRF model can be used to render previously unseen viewpoints. 

The NeRF scene representation is a pair of multilayer perceptrons (MLPs). The first MLP $f_\sigma$ takes in a 3D position $\mathbf x$ and outputs volume density $\sigma$ and a feature vector. This feature vector is concatenated with a 2D viewing direction $\mathbf d$ and fed into the second MLP $f_c$, which outputs an RGB color $\mathbf c$. This architecture ensures that the output color can vary when observed from different angles, allowing NeRF to represent reflections and glossy materials, but that the underlying geometry represented by $\sigma$ is only a function of position. 

Each pixel in an image corresponds to a ray $\mathbf r(t) = \mathbf o + t \mathbf d$ through 3D space. To calculate the color of $\mathbf r$, NeRF randomly samples distances $\{t_i\}_{i=0}^N$ along the ray and passes the points $\mathbf r(t_i)$ and direction $\mathbf d$ through its MLPs to calculate $\sigma_i$ and $ \mathbf c_i$. The resulting output color is

\begin{align}
    \mathbf c_{\textrm{out}} &= \sum_{i=1}^N w_i \mathbf c_i, 
    \quad \textrm{where }  w_i = T_i (1-e^{-\Delta_i \sigma_i}), \\
    T_i &= \exp\left( -\sum_{j < i} \Delta_j \sigma_j \right), \quad \Delta_i = t_i - t_{i-1} \,. \label{eqn:transmittance}
\end{align}
The full implementation of NeRF iteratively resamples the points $t_i$ (by treating the weights $w_i$ as a probability distribution) in order to better concentrate samples in areas of high density.

To enable the NeRF MLPs to represent higher frequency detail~\cite{tancik2020fourfeat}, the inputs $\mathbf x$ and $\mathbf d$ are each preprocessed by a componentwise sinusoidal positional encoding $\gamma_{\textrm{PE}}$:
\begin{align}
\resizebox{.9\linewidth}{!}{$
    \gamma_{\textrm{PE}}(z) = [
    \sin(2^0 z), \cos(2^0 z), \ldots, \sin(2^{L-1} z), \cos(2^{L-1} z)
    ]
    $}
\end{align}
where $L$ is the number of levels of positional encoding. 

NeRF's MLP $f_\sigma$ takes a single 3D point as input. However, this ignores both the relative footprint of the corresponding image pixel and the length of the interval $[t_{i-1}, t_i]$ along the ray $\mathbf r$ containing the point, resulting in aliasing artifacts when rendering novel camera trajectories.
Mip-NeRF~\cite{barron2021mip} remedies this issue by using the projected pixel footprint to sample conical frustums along the ray rather than intervals. To feed these frustums into the MLP, mip-NeRF approximates each of them as Gaussian distributions with parameters $\boldsymbol \mu_i, \boldsymbol \Sigma_i$ and replaces the positional encoding $\gamma_{\textrm{PE}}$ with its expectation over the input Gaussian
\begin{align}
    \gamma_{\textrm{IPE}}(\boldsymbol \mu, \boldsymbol \Sigma) = \mathbb E_{\boldsymbol X \sim \mathcal N(\boldsymbol{\mu}, \boldsymbol \Sigma)}[\gamma_{\textrm{PE}}(\boldsymbol X)] \, ,
\end{align}
referred to as an \emph{integrated} positional encoding.

\cutsectionup
\section{Method}

Training a single NeRF does not scale when trying to represent scenes as large as cities. We instead propose splitting the environment into a set of Block-NeRFs that can be independently trained in parallel and composited during inference. This independence enables the ability to expand the environment with additional Block-NeRFs or update blocks without retraining the entire environment (see Figure~\ref{fig:teaser}). We dynamically select relevant Block-NeRFs for rendering, which are then composited in a smooth manner when traversing the scene. To aid with this compositing, we optimize the appearances codes to match lighting conditions and use interpolation weights computed based on each Block-NeRF's distance to the novel view.
\cutsubsectionup
\subsection{Block Size and Placement}
\cutsubsectiondown
The individual Block-NeRFs should be arranged to collectively ensure full coverage of the target environment. We typically place one Block-NeRF at each intersection, covering the intersection itself and any connected street 75\% of the way until it converges into the next intersection (see Figure~\ref{fig:teaser}). This results in a 50\% overlap between any two adjacent blocks on the connecting street segment, making appearance alignment easier between them. Following this procedure means that the block size is variable; where necessary, additional blocks may be introduced as connectors between intersections. We ensure that the training data for each block stays exactly within its intended bounds by applying a geographical filter. This procedure can be automated and only relies on basic map data such as OpenStreetMap~\cite{haklay2008openstreetmap}.

Note that other placement heuristics are also possible, as long as the entire environment is covered by at least one Block-NeRF. For example, for some of our experiments, we instead place blocks along a single street segment at uniform distances and define the block size as a sphere around the Block-NeRF Origin (see Figure \ref{fig:merge}).
\cutsubsectionup
\subsection{Training Individual Block-NeRFs}
\cutsubsectiondown
\subsubsection{Appearance Embeddings}
\cutsubsectiondown

 \begin{figure*}
     \centering
     \includegraphics[width=1\linewidth]{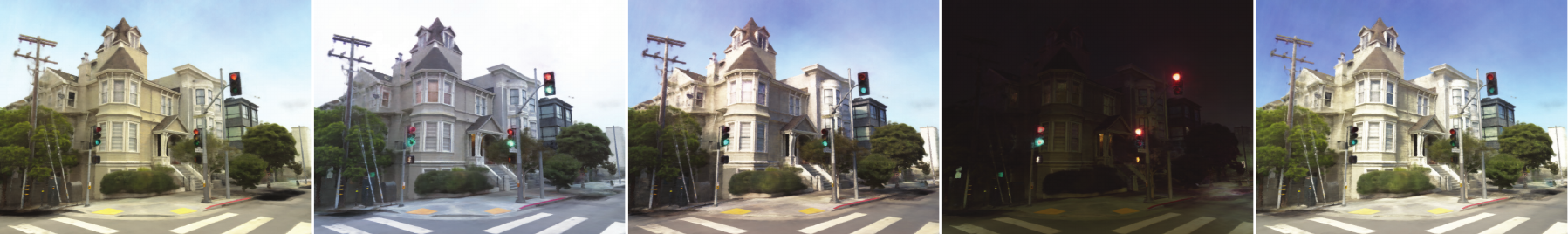}
     \cuthalfcaptionup
     \caption{The appearance codes allow the model to represent different lighting and weather conditions.}
     \cuthalfcaptionup
     \label{fig:appearance}
 \end{figure*}

Given that different parts of our data may be captured under different environmental conditions, we follow \mbox{NeRF-W}~\cite{martin2021nerf} and use Generative Latent Optimization~\cite{bojanowski2017optimizing} to optimize per-image appearance embedding vectors, as shown in Figure~\ref{fig:network}. This allows the NeRF to explain away several appearance-changing conditions, such as varying weather and lighting. We can additionally manipulate these appearance embeddings to interpolate between different conditions observed in the training data (such as cloudy versus clear skies, or day and night). Examples of rendering with different appearances can be seen in Figure~\ref{fig:appearance}. In \S~\ref{sec:appearance_matching}, we use test-time optimization over these embeddings to match the appearance of adjacent Block-NeRFs, which is important when combining multiple renderings. 

\cutsubsectionup
\subsubsection{Learned Pose Refinement}
\cutsubsectiondown
Although we assume that camera poses are provided, we find it advantageous to learn regularized pose offsets for further alignment. Pose refinement has been explored in previous NeRF based models~\cite{su2021nerf,lin2021barf,wang2021nerf,yen2020inerf}. These offsets are learned per driving segment and include both a translation and a rotation component. We optimize these offsets jointly with the NeRF itself, significantly regularizing the offsets in the early phase of training to allow the network to first learn a rough structure prior to modifying the poses.

\cutsubsectionup
\subsubsection{Exposure Input}
\cutsubsectiondown

\begin{figure}
    \centering
    \captionsetup[subfigure]{labelformat=empty}
    \begin{subfigure}{.33\linewidth}
      \centering
      \includegraphics[width=\linewidth]{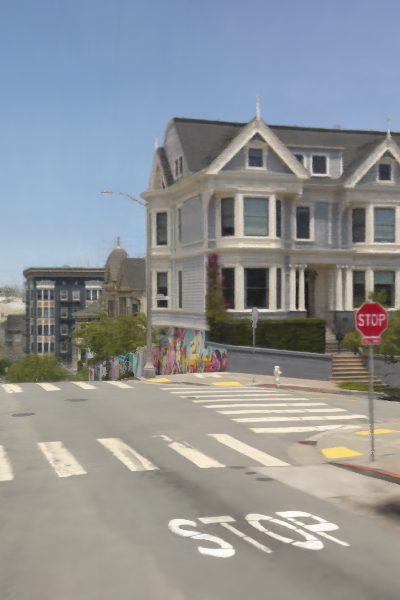}
    \end{subfigure}%
    \begin{subfigure}{.33\linewidth}
      \centering
      \includegraphics[width=\linewidth]{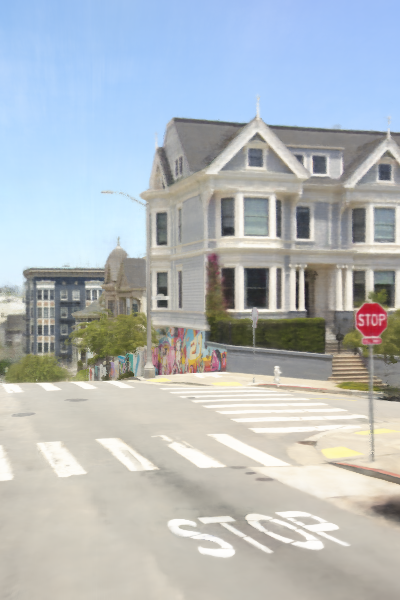}
    \end{subfigure}%
    \begin{subfigure}{.33\linewidth}
      \centering
      \includegraphics[width=\linewidth]{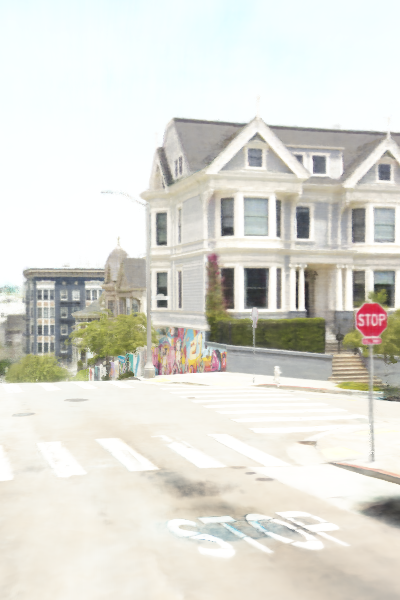}
    \end{subfigure}
    \caption{Our model is conditioned on exposure, which helps account for exposure changes present in the training data. This allows users to alter the appearance of the output images in a human-interpretable manner during inference.}
    \cuthalfcaptionup
    \label{fig:exposure}
\end{figure}

Training images may be captured across a wide range of exposure levels, which can impact NeRF training if left unaccounted for. We find that feeding the camera exposure information to the appearance prediction part of the model allows the NeRF to compensate for the visual differences (see Figure~\ref{fig:network}). Specifically, the exposure information is processed as
$\gamma_{\textrm{PE}}(\text{shutter speed} \times  \text{analog gain}/t)$
where $\gamma_{\textrm{PE}}$ is a sinusoidal positional encoding with 4 levels, and $t$ is a scaling factor (we use \SI{1000} in practice).
An example of different learned exposures can be found in Figure~\ref{fig:exposure}.

\cutsubsectionup
\subsubsection{Transient Objects}
\cutsubsectiondown
\label{sec:transient_objects}

While our method accounts for variation in appearance using the appearance embeddings, we assume that the scene geometry is consistent across the training data. Any movable objects (\eg cars, pedestrians) typically violate this assumption. We therefore use a semantic segmentation model~\cite{cheng2020panoptic} to produce masks of common movable objects, and ignore masked areas during training.
While this does not account for changes in otherwise static parts of the environment, \eg construction, it accommodates most common types of geometric inconsistency.

\cutsubsectionup
\subsubsection{Visibility Prediction}
\cutsubsectiondown
When merging multiple Block-NeRFs, it can be useful to know whether a specific region of space was visible to a given NeRF during training. We extend our model with an additional small MLP $f_v$ that is trained to learn an approximation of the \textit{visibility} of a sampled point (see Figure~\ref{fig:network}). For each sample along a training ray, $f_v$ takes in the location and view direction and regresses the corresponding transmittance of the point ($T_i$ in Equation~\ref{eqn:transmittance}). The model is trained alongside $f_{\sigma}$, which provides supervision. 
Transmittance represents how visible a point is from a particular input camera: points in free space or on the surface of the first intersected object will have transmittance near 1, and points inside or behind the first visible object will have transmittance near 0. If a point is seen from some viewpoints but not others, the regressed transmittance value will be the average over all training cameras and lie between zero and one, indicating that the point is partially observed. Our visibility prediction is similar to the visibility fields proposed by Srinivasan \etal~\cite{srinivasan2021nerv}. However, they used an MLP to predict visibility to environment lighting for the purpose of recovering a relightable NeRF model, while we predict visibility to training rays.

The visibility network is small and can be run independently from the color and density networks. This proves useful when merging multiple NeRFs, since it can help to determine whether a specific NeRF is likely to produce meaningful outputs for a given location, as explained in \S~\ref{sec:nerf_selection}. The visibility predictions can also be used to determine locations to perform appearance matching between two NeRFs, as detailed in \S~\ref{sec:appearance_matching}.

\cutsubsectionup
\subsection{Merging Multiple Block-NeRFs}
\cutsubsectiondown

\subsubsection{Block-NeRF Selection}
\cutsubsectiondown
\label{sec:nerf_selection}
The environment can be composed of an arbitrary number of Block-NeRFs. For efficiency, we utilize two filtering mechanisms to only render relevant blocks for the given target viewpoint. We only consider Block-NeRFs that are within a set radius of the target viewpoint. Additionally, for each of these candidates, we compute the associated visibility. If the mean visibility is below a threshold, we discard the Block-NeRF. An example of visibility filtering is provided in Figure~\ref{fig:merge}. Visibility can be computed quickly because its network is independent of the color network, and it does not need to be rendered at the target image resolution. After filtering, there are typically one to three Block-NeRFs left to merge.

\cutsubsectionup
\subsubsection{Block-NeRF Compositing}
\cutsubsectiondown
\label{sec:compositing}
We render color images from each of the filtered Block-NeRFs and interpolate between them using inverse distance weighting between the camera origin~$c$ and the centers~$x_i$ of each Block-NeRF. Specifically, we calculate the respective weights as $w_i\propto\operatorname{distance}(c,x_i)^{-p}$, where $p$ influences the rate of blending between Block-NeRF renders. The interpolation is done in 2D image space and produces smooth transitions between Block-NeRFs. We also explore other interpolation methods in \S~\ref{sec:exp_interp_methods}.

\cutsubsectionup
\subsubsection{Appearance Matching}
\cutsubsectiondown
\label{sec:appearance_matching}

 \begin{figure*}
     \centering
     \includegraphics[width=1\linewidth]{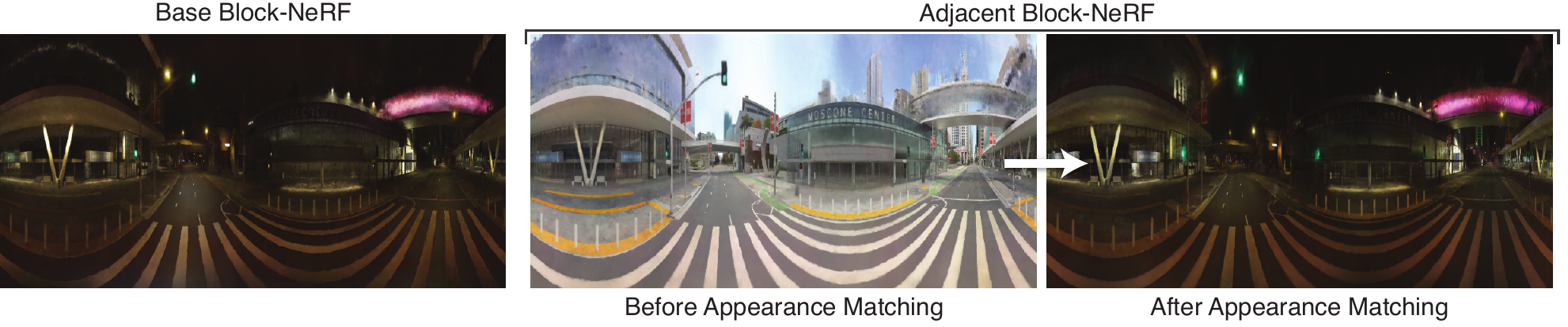}
     \caption{When rendering scenes based on multiple Block-NeRFs, we use appearance matching to obtain a consistent appearance across the scene. Given a fixed target appearance for one of the Block-NeRFs (left image), we optimize the appearances of the adjacent Block-NeRFs to match. In this example, appearance matching produces a consistent night appearance across Block-NeRFs.}
     \label{fig:matching}
 \end{figure*}

The appearance of our learned models can be controlled by an appearance latent code after the Block-NeRF has been trained. These codes are randomly initialized during training and therefore the same code typically leads to different appearances when fed into different Block-NeRFs. This is undesirable when compositing as it may lead to inconsistencies between views. Given a target appearance in one of the Block-NeRFs, we aim to match its appearance in the remaining blocks. To accomplish this, we first select a 3D \textit{matching location} between pairs of adjacent Block-NeRFs. The visibility prediction at this location should be high for both Block-NeRFs.

Given the matching location, we freeze the Block-NeRF network weights and only optimize the appearance code of the target in order to reduce the $\ell_2$ loss between the respective area renders. This optimization is quick, converging within $100$ iterations. While not necessarily yielding perfect alignment, this procedure aligns most global and low-frequency attributes of the scene, such as time of day, color balance, and weather, which is a prerequisite for successful compositing. Figure~\ref{fig:matching} shows an example optimization, where appearance matching turns a daytime scene into nighttime to match the adjacent Block-NeRF.

The optimized appearance is iteratively propagated through the scene. Starting from one root Block-NeRF, we optimize the appearance of the neighboring ones and continue the process from there. If multiple blocks surrounding a target Block-NeRF have already been optimized, we consider each of them when computing the loss.

\cutsectionup
\section{Results and Experiments}

In this section we will discuss our datasets and experiments. The architectural and optimization specifics are provided in the supplement. The supplement also provides comparisons to reconstructions from COLMAP~\cite{schonberger2016structure}, a traditional Structure from Motion approach. This reconstruction is sparse and fails to represent reflective surfaces and the sky.

\cutsubsectionup
\subsection{Datasets}
\cutsubsectiondown
We perform experiments on datasets that we collect specifically for the task of novel view synthesis of large-scale scenes. Our dataset is collected on public roads using data collection vehicles. 
While several large-scale driving datasets already exist, they are not designed for the task of view synthesis. For example, some datasets lack sufficient camera coverage (\eg, KITTI~\cite{geiger2012we}, Cityscapes~\cite{cordts2016cityscapes}) or prioritize visual diversity over repeated observations of a target area (\eg, NuScenes~\cite{caesar2020nuscenes}, Waymo Open Dataset~\cite{sun2020scalability}, Argoverse~\cite{chang2019argoverse}). Instead, they are typically designed for tasks such as object detection or tracking, where similar observations across drives can lead to generalization issues.

We capture both long-term sequence data (\SI{100}{\second} or more), as well as distinct sequences captured repeatedly in a particular target area over a period of several months. 
We use image data captured from \SI{12} cameras that collectively provide a \SI{360}{\degree}~view. \SI{8} of the cameras provide a complete surround view from the roof of the car, with \SI{4} additional cameras located at the vehicle front pointing forward and sideways. Each camera captures images at \SI{10}{\Hz} and stores a scalar exposure value. 
The vehicle pose is known and all cameras are calibrated. Using this information, we calculate the corresponding camera ray origins and directions in a common coordinate system, also accounting for the rolling shutter of the cameras. As described in \S~\ref{sec:transient_objects}, we use a semantic segmentation model~\cite{cheng2020panoptic} to detect movable objects.

\cutparagraphup
\paragraph{San Francisco Alamo Square Dataset.}  We select San Francisco's Alamo Square neighborhood as the target area for our scalability experiments. The dataset spans an area of approximately $\SI{960}{\meter}\times \SI{570}{\meter}$, and was recorded in June, July, and August of 2021. We divide this dataset into $35$ Block-NeRFs. Example renderings and Block-NeRF placements can be seen in Figure~\ref{fig:teaser}. To best appreciate the scale of the reconstruction, please refer to supplementary videos. Each Block-NeRF was trained on data from \numrange{38}{48} different data collection runs, adding up to a total driving time of \numrange{18}{28} minutes each. After filtering out some redundant image captures (\eg stationary captures), each Block-NeRF is trained on between \numrange{64575}{108216} images. The overall dataset is composed of \SI{13.4}{\hour} of driving time sourced from \SI{1330} different data collection runs, with a total of \SI{2818745} training images.

\cutparagraphup
\paragraph{San Francisco Mission Bay Dataset.} We choose San Francisco's Mission Bay District as the target area for our baseline, block size, and placement experiments. Mission Bay is an urban environment with challenging geometry and reflective facades.
We identified a long stretch on Third Street with far-range visibility, making it an interesting test case. 
Notably, this dataset was recorded in a single capture in November 2020, with consistent environmental conditions allowing for simple evaluation.
This dataset was recorded over \SI{100}{\second}, in which the data collection vehicle traveled \SI{1.08}{\km} and captured \SI{12000}~total images from \SI{12}~cameras.
We will release this single-capture dataset to aid reproducibility.  

\cutsubsectionup
\subsection{Model Ablations}
\label{sec:model_ablations}
\cutsubsectiondown
\begin{table}[!htbp]
    \centering
    \scalebox{0.75}{
    \begin{tabular}{cc||r|r|r}
    \toprule 
    & NeRFs & PSNR$\uparrow$ & SSIM$\uparrow$ & LPIPS$\downarrow$ \\
    \midrule  \midrule
    & mip-NeRF & 17.86 & 0.563 & 0.509 \\
    \midrule
    \parbox[t]{2mm}{\multirow{4}{*}{\rotatebox[origin=c]{90}{Ours}}} & -Appearance & 20.13 & 0.611 & 0.458 \\
    & -Exposure & 23.55 & \textbf{0.649} & 0.418 \\
    & -Pose Opt. & 23.05 & 0.625 & 0.442 \\
    & Full & \textbf{23.60} & \textbf{0.649} & \textbf{0.417} \\
    
 \bottomrule
 \end{tabular}
 }
    \cuthalfcaptionup
    \caption{Ablations of different Block-NeRF components on a single intersection in the Alamo Square dataset. We show the performance of mip-NeRF as a baseline, as well as the effect of removing individual components from our method.}
    \cuthalfcaptionup
    \cuthalfcaptionup
    \label{tab:model_ablation}
\end{table}
 \begin{figure*}
     \centering
     \includegraphics[width=.76\linewidth]{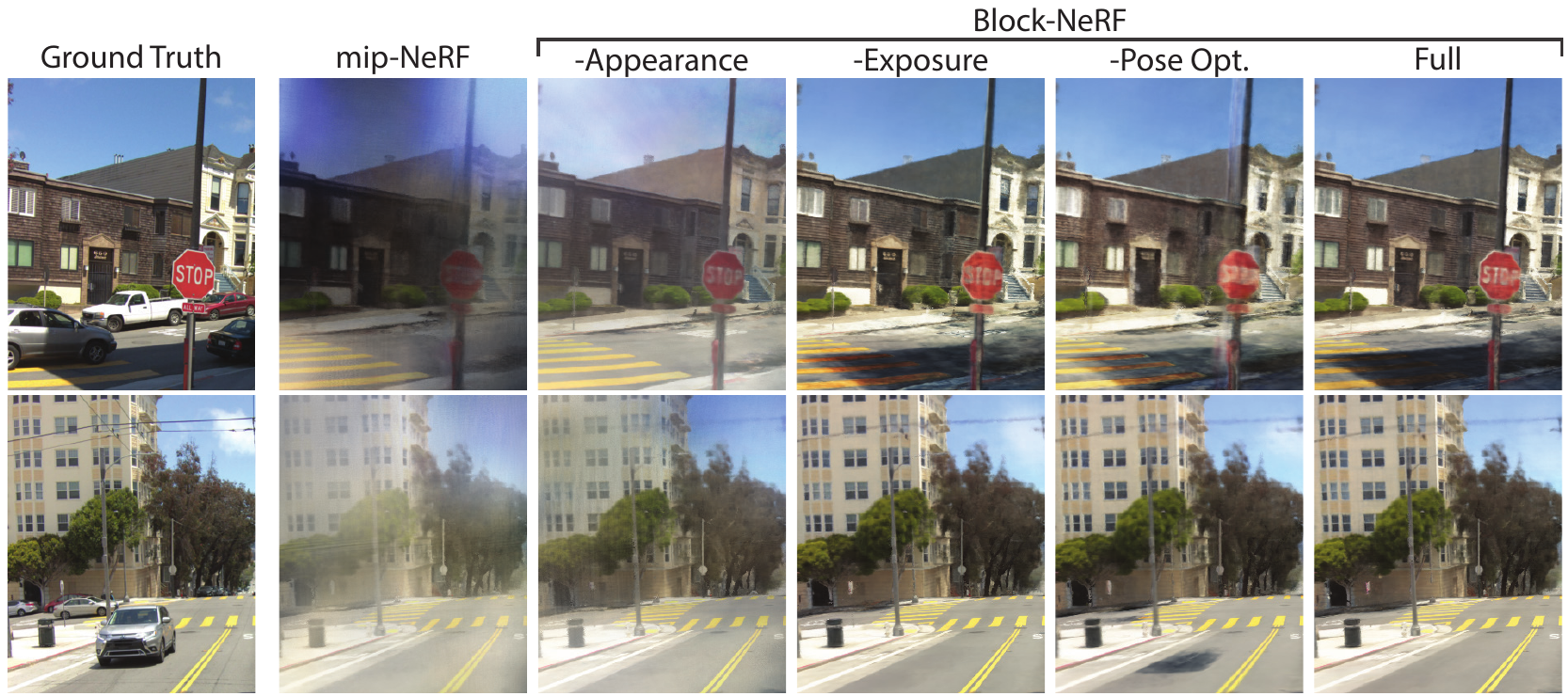}
     \caption{Model ablation results on multi segment data. Appearance embeddings help the network avoid adding cloudy geometry to explain away changes in the environment like weather and lighting. Removing exposure slightly decreases the accuracy. The pose optimization helps sharpen the results and removes ghosting from repeated objects, as observed with the telephone pole in the first row.}
     \cuthalfcaptionup
     \label{fig:ablations}
 \end{figure*}

We ablate our model modifications on a single intersection from the Alamo Square dataset. We report PSNR, SSIM, and LPIPS~\cite{zhang2018unreasonable} metrics for the test image reconstructions in Table~\ref{tab:model_ablation}. The test images are split in half vertically, with the appearance embeddings being optimized on one half and tested on the other. We also provide qualitative examples in Figure~\ref{fig:ablations}. Mip-NeRF alone fails to properly reconstruct the scene and is prone to adding non-existent geometry and cloudy artifacts to explain the differences in appearance. When our method is not trained with appearance embeddings, these artifacts are still present. If our method is not trained with pose optimization, the resulting scene is blurrier and can contain duplicated objects due to pose misalignment. Finally, the exposure input marginally improves the reconstruction, but more importantly provides us with the ability to change the exposure during inference.

\cutsubsectionup
\subsection{Block-NeRF Size and Placement}
\cutsubsectiondown

\begin{table}[!htbp]
    \centering
    \scalebox{0.68}{
    \begin{tabular}{c|r|r|r||r|r|r}
    \toprule
    \# Blocks & Weights / Total & Size & Compute & PSNR$\uparrow$ & SSIM$\uparrow$ & LPIPS$\downarrow$\\
    \midrule
    \midrule
    1 & 0.25M / 0.25M & \SI{544}{\meter} & $1\times$ & 23.83 & 0.825 & 0.381 \\
    4 & 0.25M / 1.00M & \SI{271}{\meter} & $2\times$ & 25.55 & 0.868 & 0.318  \\
    8 & 0.25M / 2.00M & \SI{116}{\meter} & $2\times$ & 26.59 & 0.890 & 0.278 \\
    16 & 0.25M / 4.00M & \SI{54}{\meter} & $2\times$ & \textbf{27.40} & \textbf{0.907} & \textbf{0.242}  \\
    \midrule
    1 & 1.00M / 1.00M & \SI{544}{\meter} & $1\times$ & 24.90 & 0.852 & 0.340 \\
    4 & 0.25M / 1.00M & \SI{271}{\meter} & $0.5\times$ & 25.55 & 0.868 & 0.318 \\
    8 & 0.13M / 1.00M & \SI{116}{\meter} & $0.25\times$ & 25.92 & 0.875 & 0.306 \\
    16 & 0.07M / 1.00M & \SI{54}{\meter} & $0.125\times$ & \textbf{25.98} & \textbf{0.877} & \textbf{0.305}\\
 \bottomrule
 \end{tabular}
 }
    \cuthalfcaptionup
    \caption{Comparison of different numbers of Block-NeRFs for reconstructing the Mission Bay dataset. Splitting the scene into multiple Block-NeRFs improves the reconstruction accuracy, even when holding the total number of weights constant (bottom section). The number of blocks determines the size of the area each block is trained on and the relative compute expense at inference time.}
    \cuthalfcaptionup
    \cuthalfcaptionup
    \label{tab:size_comparisons}
\end{table}

We compare performance on our Mission Bay dataset versus the number of Block-NeRFs used. 
We show details in Table~\ref{tab:size_comparisons}, where depending on granularity, the Block-NeRF sizes range from as small as \SI{54}{\meter} to as large as \SI{544}{\meter}. We ensure that each pair of adjacent blocks overlaps by 50\% and compare other overlap percentages in the supplement. All were evaluated on the same set of held-out test images spanning the entire trajectory. We consider two regimes, one where each Block-NeRF contains the same number of weights (top section) and one where the total number of weights across all Block-NeRFs is fixed (bottom section). In both cases, we observe that increasing the number of models improves the reconstruction metrics. In terms of computational expense, parallelization during training is trivial as each model can be optimized independently across devices. At inference, our method only requires rendering Block-NeRFs near the target view. Depending on the scene and NeRF layout, we typically render between one to three NeRFs. We report the relative compute expense in each setting without assuming any parallelization, which however would be possible and lead to an additional speed-up. Our results imply that splitting the scene into multiple lower capacity models can reduce the overall computational cost as not all of the models need to be evaluated (see bottom section of Table~\ref{tab:size_comparisons}).

\cutsubsectionup
\subsection{Interpolation Methods}
\label{sec:exp_interp_methods}
\cutsectiondown
\begin{table}[!htbp]
    \centering
    \scalebox{0.7}{
    \begin{tabular}{r||c|r|r|r}
    \toprule 
    Interpolation & Consistent? & PSNR$\uparrow$ & SSIM$\uparrow$ & LPIPS$\downarrow$ \\
    \midrule  \midrule
    Nearest & -- & 26.40 & 0.887 & 0.280 \\
    IDW 2D & \cmark & 26.59 & 0.890 & 0.278 \\
    IDW 3D & -- & 26.57 & 0.890 & 0.278 \\
    Pixelwise Visibility & -- & 27.39 & 0.906 & 0.242 \\
    Imagewise Visibility & -- & 27.41 & 0.907 & 0.242 \\
 \bottomrule
 \end{tabular}
 }
    \cuthalfcaptionup
    \caption{Comparison of interpolation methods. For our flythrough video results, we opt for 2D inverse distance weighting (IDW) as it produces temporally consistent results.}
    \cuthalfcaptionup
    \cuthalfcaptionup
    \label{tab:interpolation_comparisions}
\end{table}
We explore different interpolation methods in Table~\ref{tab:interpolation_comparisions}. The simple method of only rendering the nearest Block-NeRF to the camera requires the least amount of compute but results in harsh jumps when transitioning between blocks. These transitions can be smoothed by using inverse distance weighting (IDW) between the camera and Block-NeRF centers, as described in \S~\ref{sec:compositing}. We also explored a variant of IDW where the interpolation was performed over projected 3D points predicted by the expected Block-NeRF depth. This method suffers when the depth prediction is incorrect, leading to artifacts and temporal incoherence.

Finally, we experiment with weighing the Block-NeRFs based on per-pixel and per-image predicted visibility. This produces sharper reconstructions of further-away areas but is prone to temporal inconsistency. Therefore, these methods are best used only when rendering still images. Further details are provided in the supplement.

\cutsectionup
\section{Limitations and Future Work}
\cutsectiondown
The proposed method handles transient objects by filtering them out during training via masking using a segmentation algorithm. If objects are not properly masked, they can cause artifacts in the resulting renderings. For example, the shadows of cars often remain, even when the car itself is correctly removed. Vegetation also breaks this assumption as foliage changes seasonally and moves in the wind; this results in blurred representations of trees and plants. Similarly, temporal inconsistencies in the training data, such as construction work, are not automatically handled and require the manual retraining of the affected blocks. Further, the inability to render scenes containing dynamic objects currently limits the applicability of Block-NeRF towards closed-loop simulation tasks in robotics. In the future, these issues could be addressed by learning transient objects during the optimization~\cite{martin2021nerf}, or directly modeling dynamic objects~\cite{yang2021learning,ost2021neural}. In particular, the scene could be composed of multiple Block-NeRFs of the environment and individual controllable object NeRFs. Separation can be facilitated by the use of segmentation masks or bounding boxes.

In our model, distant objects in the scene are not sampled with the same density as nearby objects which leads to blurrier reconstructions. This is an issue with sampling unbounded volumetric representations. Techniques proposed in NeRF++~\cite{zhang2020nerf++} and concurrent Mip-NeRF 360~\cite{barron2021mip360} could potentially be used to produce sharper renderings of distant objects.

In many applications, real-time rendering is key, but NeRFs are computationally expensive to render (up to multiple seconds per image). Several NeRF caching techniques~\cite{garbin2021fastnerf,yu2021plenoctrees,hedman2021baking} or a sparse voxel grid~\cite{liu2020nsvf} could be used to enable real-time Block-NeRF rendering. Similarly, multiple concurrent works have demonstrated techniques to speed up training of NeRF style representations by multiple orders of magnitude~\cite{mueller2022instant,sun2021direct,yu2021plenoxels}.

\cutsectionup
\section{Conclusion}
\cutsectiondown
In this paper we propose Block-NeRF, a method that reconstructs arbitrarily large environments using NeRFs. We demonstrate the method's efficacy by building an entire neighborhood in San Francisco from 2.8M images, forming the largest neural scene representation to date. We accomplish this scale by splitting our representation into multiple blocks that can be optimized independently. At such a scale, the data collected will necessarily have transient objects and variations in appearance, which we account for by modifying the underlying NeRF architecture. We hope that this can inspire future work in large-scale scene reconstruction using modern neural rendering methods.

{\small
\bibliographystyle{ieee_fullname}
\bibliography{reference}

\begin{thebibliography}{10}\itemsep=-1pt

\bibitem{agarwal2011building}
Sameer Agarwal, Yasutaka Furukawa, Noah Snavely, Ian Simon, Brian Curless,
  Steven~M Seitz, and Richard Szeliski.
\newblock Building rome in a day.
\newblock {\em Communications of the ACM}, 2011.

\bibitem{amini2020learning}
Alexander Amini, Igor Gilitschenski, Jacob Phillips, Julia Moseyko, Rohan
  Banerjee, Sertac Karaman, and Daniela Rus.
\newblock Learning robust control policies for end-to-end autonomous driving
  from data-driven simulation.
\newblock {\em IEEE Robotics and Automation Letters}, 2020.

\bibitem{barron2021mip}
Jonathan~T Barron, Ben Mildenhall, Matthew Tancik, Peter Hedman, Ricardo
  Martin-Brualla, and Pratul~P Srinivasan.
\newblock Mip-{NeRF}: A multiscale representation for anti-aliasing neural
  radiance fields.
\newblock {\em ICCV}, 2021.

\bibitem{barron2021mip360}
Jonathan~T Barron, Ben Mildenhall, Dor Verbin, Pratul~P Srinivasan, and Peter
  Hedman.
\newblock Mip-nerf 360: Unbounded anti-aliased neural radiance fields.
\newblock {\em arXiv preprint arXiv:2111.12077}, 2021.

\bibitem{bojanowski2017optimizing}
Piotr Bojanowski, Armand Joulin, David Lopez-Paz, and Arthur Szlam.
\newblock Optimizing the latent space of generative networks.
\newblock {\em arXiv:1707.05776}, 2017.

\bibitem{buehler2001unstructured}
Chris Buehler, Michael Bosse, Leonard McMillan, Steven Gortler, and Michael
  Cohen.
\newblock Unstructured lumigraph rendering.
\newblock {\em Computer graphics and interactive techniques}, 2001.

\bibitem{caesar2020nuscenes}
Holger Caesar, Varun Bankiti, Alex~H Lang, Sourabh Vora, Venice~Erin Liong,
  Qiang Xu, Anush Krishnan, Yu Pan, Giancarlo Baldan, and Oscar Beijbom.
\newblock nuscenes: A multimodal dataset for autonomous driving.
\newblock {\em CVPR}, 2020.

\bibitem{chang2019argoverse}
Ming-Fang Chang, John Lambert, Patsorn Sangkloy, Jagjeet Singh, Slawomir Bak,
  Andrew Hartnett, De Wang, Peter Carr, Simon Lucey, Deva Ramanan, et~al.
\newblock Argoverse: 3d tracking and forecasting with rich maps.
\newblock {\em CVPR}, 2019.

\bibitem{chen2021geosim}
Yun Chen, Frieda Rong, Shivam Duggal, Shenlong Wang, Xinchen Yan, Sivabalan
  Manivasagam, Shangjie Xue, Ersin Yumer, and Raquel Urtasun.
\newblock Geosim: Realistic video simulation via geometry-aware composition for
  self-driving.
\newblock {\em CVPR}, 2021.

\bibitem{cheng2020panoptic}
Bowen Cheng, Maxwell~D Collins, Yukun Zhu, Ting Liu, Thomas~S Huang, Hartwig
  Adam, and Liang-Chieh Chen.
\newblock Panoptic-deeplab: A simple, strong, and fast baseline for bottom-up
  panoptic segmentation.
\newblock {\em CVPR}, 2020.

\bibitem{cordts2016cityscapes}
Marius Cordts, Mohamed Omran, Sebastian Ramos, Timo Rehfeld, Markus Enzweiler,
  Rodrigo Benenson, Uwe Franke, Stefan Roth, and Bernt Schiele.
\newblock The cityscapes dataset for semantic urban scene understanding.
\newblock {\em CVPR}, 2016.

\bibitem{devaranjan2020meta}
Jeevan Devaranjan, Amlan Kar, and Sanja Fidler.
\newblock Meta-sim2: Unsupervised learning of scene structure for synthetic
  data generation.
\newblock {\em ECCV}, 2020.

\bibitem{dosovitskiy2017carla}
Alexey Dosovitskiy, German Ros, Felipe Codevilla, Antonio Lopez, and Vladlen
  Koltun.
\newblock Carla: An open urban driving simulator.
\newblock {\em Conference on robot learning}, 2017.

\bibitem{du2018unmanned}
Dawei Du, Yuankai Qi, Hongyang Yu, Yifan Yang, Kaiwen Duan, Guorong Li, Weigang
  Zhang, Qingming Huang, and Qi Tian.
\newblock The unmanned aerial vehicle benchmark: Object detection and tracking.
\newblock {\em ECCV}, 2018.

\bibitem{flynn2016deepstereo}
John Flynn, Ivan Neulander, James Philbin, and Noah Snavely.
\newblock Deepstereo: Learning to predict new views from the world's imagery.
\newblock {\em CVPR}, 2016.

\bibitem{fruh2004automated}
Christian Fr{\"u}h and Avideh Zakhor.
\newblock An automated method for large-scale, ground-based city model
  acquisition.
\newblock {\em IJCV}, 2004.

\bibitem{furukawa2010towards}
Yasutaka Furukawa, Brian Curless, Steven~M Seitz, and Richard Szeliski.
\newblock Towards internet-scale multi-view stereo.
\newblock {\em CVPR}, 2010.

\bibitem{furukawa2010pmvs}
Yasutaka Furukawa and Jean Ponce.
\newblock Accurate, dense, and robust multi-view stereopsis.
\newblock {\em IEEE TPAMI}, 2010.

\bibitem{gaidon2016virtual}
Adrien Gaidon, Qiao Wang, Yohann Cabon, and Eleonora Vig.
\newblock Virtual worlds as proxy for multi-object tracking analysis.
\newblock {\em CVPR}, 2016.

\bibitem{garbin2021fastnerf}
Stephan~J Garbin, Marek Kowalski, Matthew Johnson, Jamie Shotton, and Julien
  Valentin.
\newblock Fastnerf: High-fidelity neural rendering at 200fps.
\newblock {\em arXiv:2103.10380}, 2021.

\bibitem{geiger2012we}
Andreas Geiger, Philip Lenz, and Raquel Urtasun.
\newblock Are we ready for autonomous driving? the kitti vision benchmark
  suite.
\newblock {\em CVPR}, 2012.

\bibitem{haklay2008openstreetmap}
Mordechai Haklay and Patrick Weber.
\newblock Openstreetmap: User-generated street maps.
\newblock {\em IEEE Pervasive computing}, 2008.

\bibitem{hartleyziss2004}
R.~I. Hartley and A. Zisserman.
\newblock {\em Multiple View Geometry in Computer Vision}.
\newblock Cambridge University Press, second edition, 2004.

\bibitem{hedman2018deep}
Peter Hedman, Julien Philip, True Price, Jan-Michael Frahm, George Drettakis,
  and Gabriel Brostow.
\newblock Deep blending for free-viewpoint image-based rendering.
\newblock {\em ACM Transactions on Graphics (TOG)}, 2018.

\bibitem{hedman2021baking}
Peter Hedman, Pratul~P Srinivasan, Ben Mildenhall, Jonathan~T Barron, and Paul
  Debevec.
\newblock Baking neural radiance fields for real-time view synthesis.
\newblock {\em arXiv:2103.14645}, 2021.

\bibitem{kar2019meta}
Amlan Kar, Aayush Prakash, Ming-Yu Liu, Eric Cameracci, Justin Yuan, Matt
  Rusiniak, David Acuna, Antonio Torralba, and Sanja Fidler.
\newblock Meta-sim: Learning to generate synthetic datasets.
\newblock {\em ICCV}, 2019.

\bibitem{kazhdan2013screened}
Michael Kazhdan and Hugues Hoppe.
\newblock Screened poisson surface reconstruction.
\newblock {\em ACM Transactions on Graphics (ToG)}, 2013.

\bibitem{kim2021drivegan}
Seung~Wook Kim, Jonah Philion, Antonio Torralba, and Sanja Fidler.
\newblock Drivegan: Towards a controllable high-quality neural simulation.
\newblock {\em CVPR}, 2021.

\bibitem{kingma2014adam}
Diederik~P Kingma and Jimmy Ba.
\newblock Adam: A method for stochastic optimization.
\newblock {\em ICLR}, 2015.

\bibitem{kopf2010street}
Johannes Kopf, Billy Chen, Richard Szeliski, and Michael Cohen.
\newblock Street slide: browsing street level imagery.
\newblock {\em ACM Transactions on Graphics (TOG)}, 2010.

\bibitem{kopf2014hyperlapse}
Johannes Kopf, Michael Cohen, and Rick Szeliski.
\newblock First-person hyperlapse videos.
\newblock {\em SIGGRAPH}, 2014.

\bibitem{li2019aads}
Wei Li, CW Pan, Rong Zhang, JP Ren, YX Ma, Jin Fang, FL Yan, QC Geng, XY Huang,
  HJ Gong, et~al.
\newblock Aads: Augmented autonomous driving simulation using data-driven
  algorithms.
\newblock {\em Science robotics}, 2019.

\bibitem{li2008modeling}
Xiaowei Li, Changchang Wu, Christopher Zach, Svetlana Lazebnik, and Jan-Michael
  Frahm.
\newblock Modeling and recognition of landmark image collections using iconic
  scene graphs.
\newblock {\em ECCV}, 2008.

\bibitem{lin2021barf}
Chen-Hsuan Lin, Wei-Chiu Ma, Antonio Torralba, and Simon Lucey.
\newblock Barf: Bundle-adjusting neural radiance fields.
\newblock {\em arXiv preprint arXiv:2104.06405}, 2021.

\bibitem{liu2021infinite}
Andrew Liu, Richard Tucker, Varun Jampani, Ameesh Makadia, Noah Snavely, and
  Angjoo Kanazawa.
\newblock Infinite nature: Perpetual view generation of natural scenes from a
  single image.
\newblock {\em ICCV}, 2021.

\bibitem{liu2020nsvf}
Lingjie Liu, Jiatao Gu, Kyaw~Zaw Lin, Tat{-}Seng Chua, and Christian Theobalt.
\newblock Neural sparse voxel fields.
\newblock {\em NeurIPS}, 2020.

\bibitem{lombardi2019neuralvolumes}
Stephen Lombardi, Tomas Simon, Jason Saragih, Gabriel Schwartz, Andreas
  Lehrmann, and Yaser Sheikh.
\newblock Neural volumes: Learning dynamic renderable volumes from images.
\newblock {\em SIGGRAPH}, 2019.

\bibitem{losasso2004geometry}
Frank Losasso and Hugues Hoppe.
\newblock Geometry clipmaps: terrain rendering using nested regular grids.
\newblock {\em Siggraph}, 2004.

\bibitem{lowe2004sift}
David~G Lowe.
\newblock Distinctive image features from scale-invariant keypoints.
\newblock {\em IJCV}, 2004.

\bibitem{martin2021nerf}
Ricardo Martin-Brualla, Noha Radwan, Mehdi~SM Sajjadi, Jonathan~T Barron,
  Alexey Dosovitskiy, and Daniel Duckworth.
\newblock Nerf in the wild: Neural radiance fields for unconstrained photo
  collections.
\newblock {\em CVPR}, 2021.

\bibitem{meshry2019rerendering}
Moustafa Meshry, Dan~B. Goldman, Sameh Khamis, Hugues Hoppe, Rohit Pandey, Noah
  Snavely, and Ricardo Martin-Brualla.
\newblock Neural rerendering in the wild.
\newblock {\em CVPR}, 2019.

\bibitem{mildenhall2020nerf}
Ben Mildenhall, Pratul~P Srinivasan, Matthew Tancik, Jonathan~T Barron, Ravi
  Ramamoorthi, and Ren Ng.
\newblock Nerf: Representing scenes as neural radiance fields for view
  synthesis.
\newblock {\em ECCV}, 2020.

\bibitem{mueller2022instant}
Thomas M\"uller, Alex Evans, Christoph Schied, and Alexander Keller.
\newblock Instant neural graphics primitives with a multiresolution hash
  encoding.
\newblock {\em arXiv:2201.05989}, Jan. 2022.

\bibitem{ost2021neural}
Julian Ost, Fahim Mannan, Nils Thuerey, Julian Knodt, and Felix Heide.
\newblock Neural scene graphs for dynamic scenes.
\newblock {\em CVPR}, 2021.

\bibitem{park2020deformable}
Keunhong Park, Utkarsh Sinha, Jonathan~T Barron, Sofien Bouaziz, Dan~B Goldman,
  Steven~M Seitz, and Ricardo Martin-Brualla.
\newblock Nerfies: Deformable neural radiance fields.
\newblock {\em ICCV}, 2021.

\bibitem{peng2020convolutional}
Songyou Peng, Michael Niemeyer, Lars Mescheder, Marc Pollefeys, and Andreas
  Geiger.
\newblock Convolutional occupancy networks.
\newblock In {\em Computer Vision--ECCV 2020: 16th European Conference,
  Glasgow, UK, August 23--28, 2020, Proceedings, Part III 16}, pages 523--540.
  Springer, 2020.

\bibitem{pollefeys2008detailed}
Marc Pollefeys, David Nist{\'e}r, J-M Frahm, Amir Akbarzadeh, Philippos
  Mordohai, Brian Clipp, Chris Engels, David Gallup, S-J Kim, Paul Merrell,
  et~al.
\newblock Detailed real-time urban 3d reconstruction from video.
\newblock {\em IJCV}, 2008.

\bibitem{rebain2021derf}
Daniel Rebain, Wei Jiang, Soroosh Yazdani, Ke Li, Kwang~Moo Yi, and Andrea
  Tagliasacchi.
\newblock Derf: Decomposed radiance fields.
\newblock {\em CVPR}, 2021.

\bibitem{reiser2021kilonerf}
Christian Reiser, Songyou Peng, Yiyi Liao, and Andreas Geiger.
\newblock {KiloNeRF}: Speeding up neural radiance fields with thousands of tiny
  {MLP}s.
\newblock {\em ICCV}, 2021.

\bibitem{richter2021enhancing}
Stephan~R Richter, Hassan~Abu AlHaija, and Vladlen Koltun.
\newblock Enhancing photorealism enhancement.
\newblock {\em arXiv:2105.04619}, 2021.

\bibitem{richter2016playing}
Stephan~R Richter, Vibhav Vineet, Stefan Roth, and Vladlen Koltun.
\newblock Playing for data: Ground truth from computer games.
\newblock {\em ECCV}, 2016.

\bibitem{riegler2020fvs}
Gernot Riegler and Vladlen Koltun.
\newblock Free view synthesis.
\newblock {\em ECCV}, 2020.

\bibitem{riegler2021svs}
Gernot Riegler and Vladlen Koltun.
\newblock Stable view synthesis.
\newblock {\em CVPR}, 2021.

\bibitem{ros2016synthia}
German Ros, Laura Sellart, Joanna Materzynska, David Vazquez, and Antonio~M
  Lopez.
\newblock The synthia dataset: A large collection of synthetic images for
  semantic segmentation of urban scenes.
\newblock {\em CVPR}, 2016.

\bibitem{schonberger2016structure}
Johannes~L Schonberger and Jan-Michael Frahm.
\newblock Structure-from-motion revisited.
\newblock {\em CVPR}, 2016.

\bibitem{shan2013turing}
Qi Shan, Riley Adams, Brian Curless, Yasutaka Furukawa, and Steven~M. Seitz.
\newblock The visual turing test for scene reconstruction.
\newblock {\em 3DV}, 2013.

\bibitem{snavely2006phototourism}
Noah Snavely, Steven~M. Seitz, and Richard Szeliski.
\newblock Photo tourism: Exploring photo collections in 3d.
\newblock {\em SIGGRAPH}, 2006.

\bibitem{srinivasan2021nerv}
Pratul~P. Srinivasan, Boyang Deng, Xiuming Zhang, Matthew Tancik, Ben
  Mildenhall, and Jonathan~T. Barron.
\newblock {NeRV}: Neural reflectance and visibility fields for relighting and
  view synthesis.
\newblock {\em CVPR}, 2021.

\bibitem{su2021nerf}
Shih-Yang Su, Frank Yu, Michael Zollh{\"o}fer, and Helge Rhodin.
\newblock A-nerf: Articulated neural radiance fields for learning human shape,
  appearance, and pose.
\newblock {\em Advances in Neural Information Processing Systems}, 34, 2021.

\bibitem{sun2021direct}
Cheng Sun, Min Sun, and Hwann-Tzong Chen.
\newblock Direct voxel grid optimization: Super-fast convergence for radiance
  fields reconstruction.
\newblock {\em arXiv preprint arXiv:2111.11215}, 2021.

\bibitem{sun2020scalability}
Pei Sun, Henrik Kretzschmar, Xerxes Dotiwalla, Aurelien Chouard, Vijaysai
  Patnaik, Paul Tsui, James Guo, Yin Zhou, Yuning Chai, Benjamin Caine, et~al.
\newblock Scalability in perception for autonomous driving: Waymo open dataset.
\newblock {\em CVPR}, 2020.

\bibitem{takikawa2021nglod}
Towaki Takikawa, Joey Litalien, Kangxue Yin, Karsten Kreis, Charles Loop, Derek
  Nowrouzezahrai, Alec Jacobson, Morgan McGuire, and Sanja Fidler.
\newblock Neural geometric level of detail: Real-time rendering with implicit
  {3D} shapes.
\newblock {\em CVPR}, 2021.

\bibitem{tancik2020fourfeat}
Matthew Tancik, Pratul~P. Srinivasan, Ben Mildenhall, Sara Fridovich-Keil,
  Nithin Raghavan, Utkarsh Singhal, Ravi Ramamoorthi, Jonathan~T. Barron, and
  Ren Ng.
\newblock Fourier features let networks learn high frequency functions in low
  dimensional domains.
\newblock {\em NeurIPS}, 2020.

\bibitem{thrun2002probabilistic}
Sebastian Thrun.
\newblock Probabilistic robotics.
\newblock {\em Communications of the ACM}, 2002.

\bibitem{triggs1999bundle}
Bill Triggs, Philip~F McLauchlan, Richard~I Hartley, and Andrew~W Fitzgibbon.
\newblock Bundle adjustment—a modern synthesis.
\newblock {\em International workshop on vision algorithms}, 1999.

\bibitem{wang2021nerf}
Zirui Wang, Shangzhe Wu, Weidi Xie, Min Chen, and Victor~Adrian Prisacariu.
\newblock Nerf--: Neural radiance fields without known camera parameters.
\newblock {\em arXiv preprint arXiv:2102.07064}, 2021.

\bibitem{yang2021learning}
Bangbang Yang, Yinda Zhang, Yinghao Xu, Yijin Li, Han Zhou, Hujun Bao, Guofeng
  Zhang, and Zhaopeng Cui.
\newblock Learning object-compositional neural radiance field for editable
  scene rendering.
\newblock {\em ICCV}, 2021.

\bibitem{yang2020surfelgan}
Zhenpei Yang, Yuning Chai, Dragomir Anguelov, Yin Zhou, Pei Sun, Dumitru Erhan,
  Sean Rafferty, and Henrik Kretzschmar.
\newblock Surfelgan: Synthesizing realistic sensor data for autonomous driving.
\newblock {\em CVPR}, 2020.

\bibitem{yariv2020multiview}
Lior Yariv, Yoni Kasten, Dror Moran, Meirav Galun, Matan Atzmon, Basri Ronen,
  and Yaron Lipman.
\newblock Multiview neural surface reconstruction by disentangling geometry and
  appearance.
\newblock {\em Advances in Neural Information Processing Systems}, 33, 2020.

\bibitem{yen2020inerf}
Lin Yen-Chen, Pete Florence, Jonathan~T. Barron, Alberto Rodriguez, Phillip
  Isola, and Tsung-Yi Lin.
\newblock {iNeRF}: Inverting neural radiance fields for pose estimation.
\newblock In {\em IEEE/RSJ International Conference on Intelligent Robots and
  Systems ({IROS})}, 2021.

\bibitem{yu2021plenoxels}
Alex Yu, Sara Fridovich-Keil, Matthew Tancik, Qinhong Chen, Benjamin Recht, and
  Angjoo Kanazawa.
\newblock Plenoxels: Radiance fields without neural networks.
\newblock {\em arXiv preprint arXiv:2112.05131}, 2021.

\bibitem{yu2021plenoctrees}
Alex Yu, Ruilong Li, Matthew Tancik, Hao Li, Ren Ng, and Angjoo Kanazawa.
\newblock Plenoctrees for real-time rendering of neural radiance fields.
\newblock {\em arXiv:2103.14024}, 2021.

\bibitem{zhang2021editable}
Jiakai Zhang, Xinhang Liu, Xinyi Ye, Fuqiang Zhao, Yanshun Zhang, Minye Wu,
  Yingliang Zhang, Lan Xu, and Jingyi Yu.
\newblock Editable free-viewpoint video using a layered neural representation.
\newblock {\em ACM Transactions on Graphics (TOG)}, 2021.

\bibitem{zhang2020nerf++}
Kai Zhang, Gernot Riegler, Noah Snavely, and Vladlen Koltun.
\newblock Nerf++: Analyzing and improving neural radiance fields.
\newblock {\em arXiv preprint arXiv:2010.07492}, 2020.

\bibitem{zhang2018unreasonable}
Richard Zhang, Phillip Isola, Alexei~A Efros, Eli Shechtman, and Oliver Wang.
\newblock The unreasonable effectiveness of deep features as a perceptual
  metric.
\newblock {\em CVPR}, 2018.

\bibitem{zhou2018stereo}
Tinghui Zhou, Richard Tucker, John Flynn, Graham Fyffe, and Noah Snavely.
\newblock Stereo magnification: Learning view synthesis using multiplane
  images.
\newblock {\em arXiv:1805.09817}, 2018.

\bibitem{zhu2018very}
Siyu Zhu, Runze Zhang, Lei Zhou, Tianwei Shen, Tian Fang, Ping Tan, and Long
  Quan.
\newblock Very large-scale global {SFM} by distributed motion averaging.
\newblock {\em CVPR}, 2018.

\end{thebibliography}
}

\clearpage

\begin{appendices}

\section{Model Parameters / Optimization Details}

Our network follows the mip-NeRF structure. The network $f_\sigma$ is composed of 8 layers with width 512 (Mission Bay experiments) or 1024 (all other experiments). $f_c$ has 3 layers with width 128 and $f_v$ has 4 layers with width 128. The appearance embeddings are 32 dimensional. We train each Block-NeRF using the Adam~\cite{kingma2014adam} optimizer for \SI{300}K iterations with a batch size of 16384.  Similar to mip-NeRF, the learning rate is an annealed logarithmically from $2 \cdot 10^{-3}$ to $2 \cdot 10^{-5}$, with a warm up phase during the first 1024 iterations. The coarse and fine networks are sampled 256 times during training and 512 times when rendering the videos. The visibility is supervised with MSE loss and is scaled by $10^{-6}$. The learned pose correction consists of a position offset and a $3\times3$ residual rotation matrix, which is added to the identity matrix and normalized before being applied to ensure it is orthogonal. The pose corrections are initialized to 0 and their element-wise $\ell2$ norm is regularized during training. This regularization is scaled by $10^{5}$ at the start of training and linearly decays to $10^{-1}$ after 5000 iterations. This allows the network to learn initial geometry prior to applying pose offsets.

Each Block-NeRF takes between 9 and 24 hours to train (depending on hyperparameters). We train each Block-NeRF on 32 TPU v3 cores available through Google Cloud Compute, which combined offer a total of $1680$ TFLOPS and $512$ GB memory. Rendering an $1200\times 900$px image for a single Block-NeRF takes approximately $5.9$ seconds. Multiple Block-NeRF can be processed in parallel during inference (typically fewer than 3 Block-NeRFs need to be rendered for a single frame).

\section{Block-NeRF Size and Placement}
\begin{figure*}
     \centering
     \includegraphics[width=1.0\linewidth]{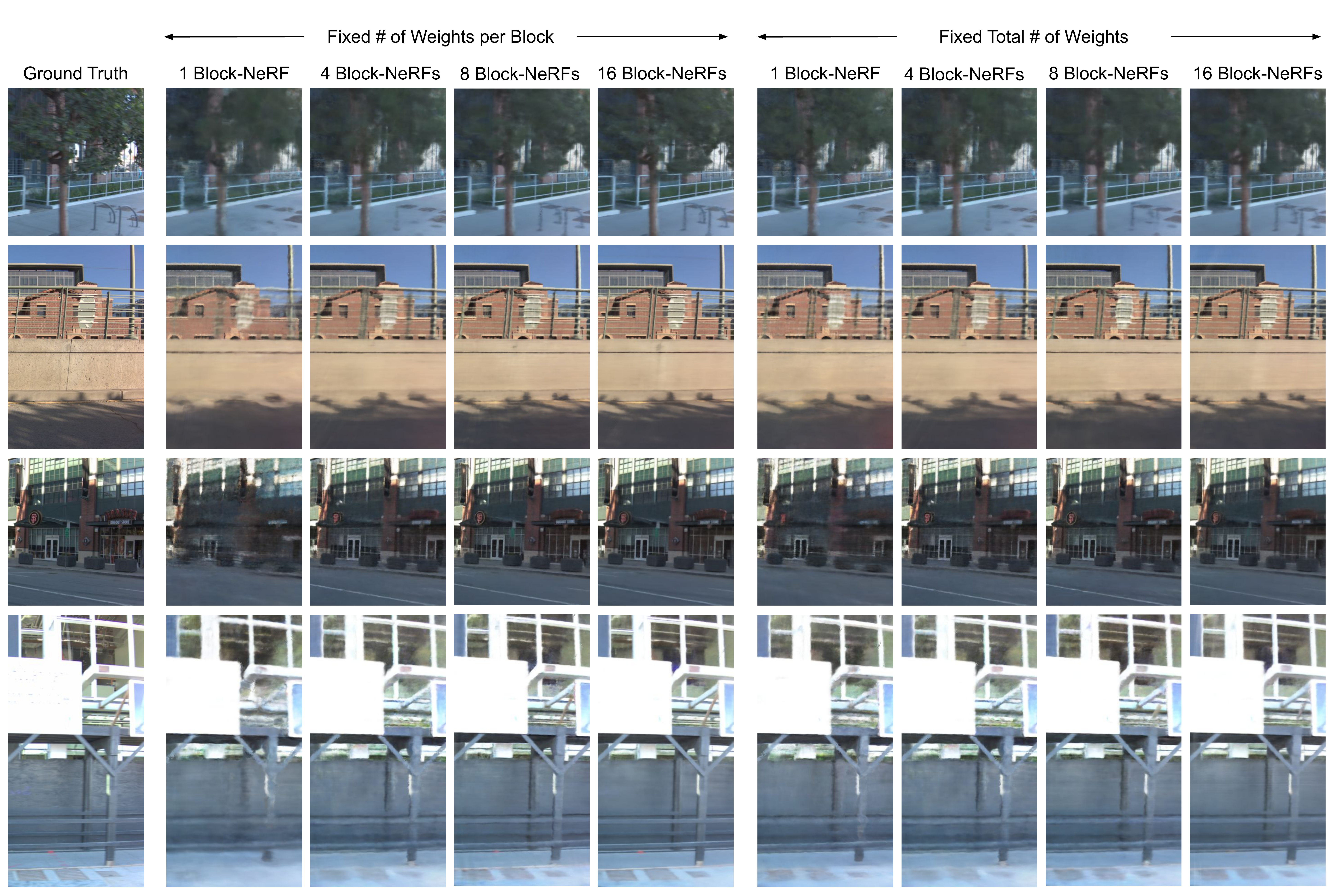}
     \caption{Qualitative results on Block-NeRF size and placement. We show results on the Mission Bay dataset using different options discussed in \S~5.3 of the main paper.  }
     \cuthalfcaptionup
     \label{fig:block_size_ablation}
 \end{figure*}

We include qualitative comparisons in Figure~\ref{fig:block_size_ablation} on the Mission Bay dataset to complement the quantitative comparisons in (\S 5.3, Table~2).
In this figure, we provide comparisons on two regimes, one where each Block-NeRF contains the same number of weights (left section) and one where the total number of weights across all Block-NeRFs is fixed (right section).

\section{Block-NeRF Overlap Comparison}

In the main paper, we include experiments on Block-NeRF size and placement (\S 5.3). For these experiments, we assumed a relative overlap of 50\% between each pair of Block-NeRFs, which aids with appearance alignment. 

Table~\ref{tab:overlap_comparisons} is a direct extension of Table 2 in the main paper and shows the effect of varying block overlap in the 8 block scenario. Note that varying the overlap changes the spatial block size. The original setting in the main paper is marked with an asterisk.

The metrics imply that reducing overlap is beneficial for image quality metrics. However, this can likely be attributed to the resulting reduction in block size. In practice, having an overlap between blocks is important to avoid temporal artifacts when interpolating between Block-NeRFs. 

\begin{table}[!htbp]
    \centering
    \scalebox{0.95}{  
    \begin{tabular}{l|r||r|r|r}
    \toprule
    Overlap & Size & PSNR$\uparrow$ & SSIM$\uparrow$ & LPIPS$\downarrow$ \\
    \midrule
    0\% & \SI{77}{\meter} & 26.77 & 0.895 & 0.262 \\
    25\% & \SI{97}{\meter} & 26.75 & 0.894 & 0.269 \\
    50\%* & \SI{116}{\meter} & 26.59 & 0.890 & 0.278 \\
    75\% & \SI{136}{\meter} & 26.51 & 0.887 & 0.283 \\
 \bottomrule
 \end{tabular}
 }
    \cuthalfcaptionup
    \caption{Effect of different NeRF overlaps in the 8 block scenario with 0.25M weights per block (2M weights in total). The original setting used in the main paper is marked*.}
    \cuthalfcaptionup
    \cuthalfcaptionup
    \label{tab:overlap_comparisons}
\end{table}

\section{Block-NeRF Interpolation Details}

We experiment with multiple methods to interpolate between Block-NeRFs and find that simple inverse distance weighting (IDW) in image space produces the most appealing videos due to temporal smoothness. We use an IDW power $p$ of 4 for the Alamo Square renderings and a power of 1 for the Mission Bay renderings. We experiment with 3D inverse distance weighting for each individual pixel by projecting the rendered pixels into 3D space using the expected ray termination depth from the Block-NeRF closest to the target view. The color value of the projected pixel is then determined using inverse distance weighting with the nearest Block-NeRFs. Artifacts occur in the resulting composited renders due to noise in the depth predictions. We also experiment with using the Block-NeRF predicted visibility for interpolation. We consider imagewise visibility where we take the mean visibility of the entire image and pixelwise visibility where were directly utilize the per-pixel visibility predictions. Both of these methods lead to sharper results but come at the cost of temporal inconsistencies. Finally we compare to nearest neighbor interpolation where we only render the Block-NeRF closest to the target view. This results in harsh jumps when transiting between Block-NeRFs.

\section{Structure from Motion (COLMAP)}

We use COLMAP~\cite{schonberger2016structure} to reconstruct the Mission Bay dataset. We first split the dataset into $8$ overlapping blocks with \SI{97}{\meter} radius each based on camera positions (each block has roughly 25\% overlap with the adjacent block).
The bundle adjustment step takes most of the time in reconstruction and we do not see significant improvements if we increase the radius per block.
We mask out movable objects when extracting feature points for matching, using the same segmentation model as Block-NeRF. We assume a pinhole camera model and provide camera intrinsics and camera pose as priors for running structure-from-motion. 
We then run multi-view stereo within each block to produce dense depth and normal maps in 3D and produce a dense point cloud of the scene.
In our preliminary experiments, we ran Poisson meshing~\cite{kazhdan2013screened} on the fused dense pointcloud to reconstruct textured meshes but found that the method fails to produce reasonably-looking results due to the challenging geometry and depth errors introduced by reflective surfaces and the sky.
Instead, we leverage the fused pointcloud and explore two alternatives, namely, point rendering and surfel rendering, respectively.
To render the test view, we selected the nearest scene and use \texttt{OSMesa}\footnote{\url{https://docs.mesa3d.org/osmesa.html}} off-screen rendering assuming the Lambertian model and a single light source.

In Table~\ref{tab:colmap_comparisons}, we compare two different rendering options for the densely reconstructed pointcloud.
We discard the invisible pixels when computing the PSNR for both methods, making the quantitative results comparable to our Block-NeRF setting.

\begin{figure}
     \centering
     \includegraphics[width=1.0\linewidth]{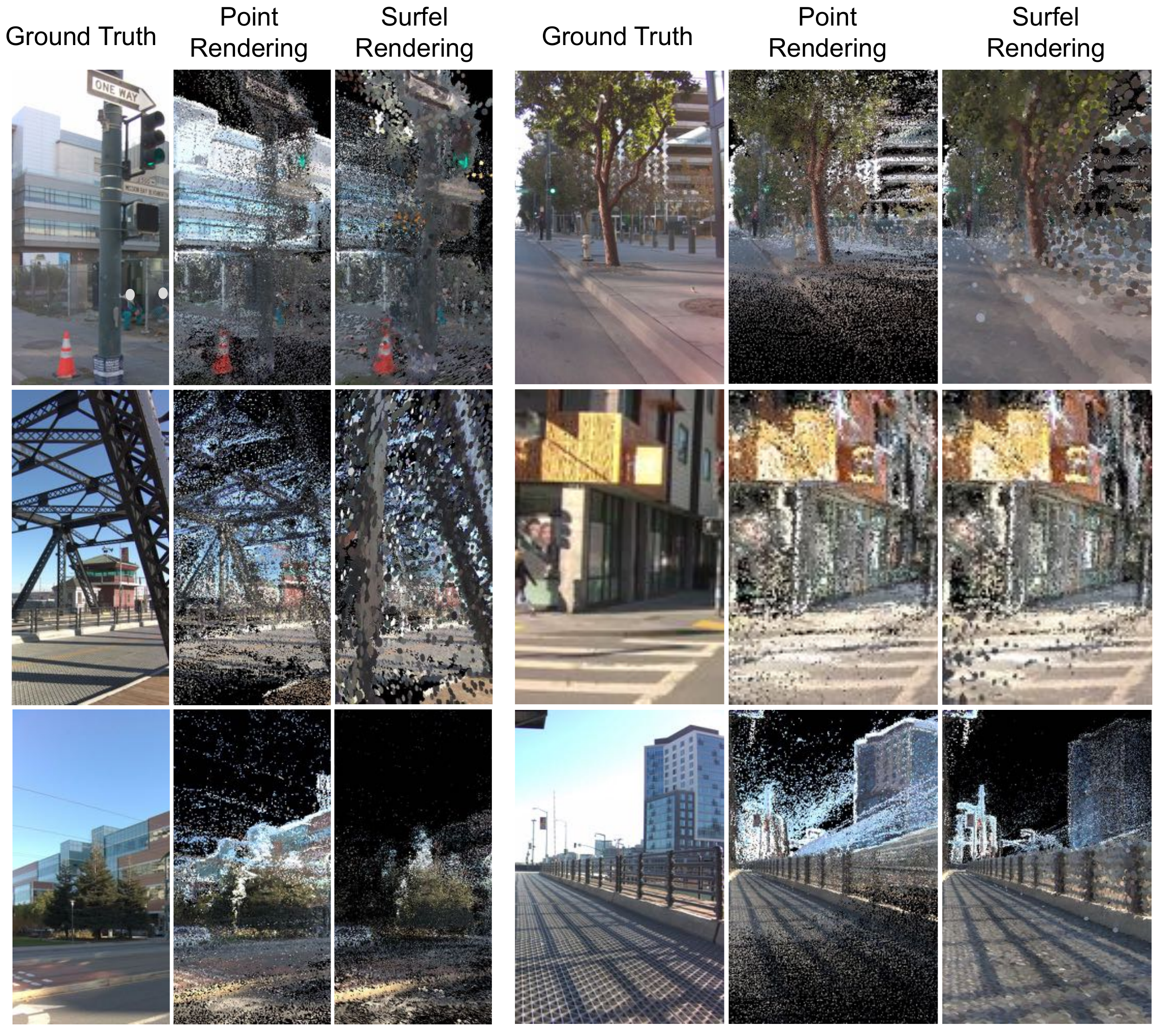}
     \caption{Qualitative results for COLMAP. We demonstrate the two rendering options using the fused pointcloud computed by COLMAP.}
     \cuthalfcaptionup
     \label{fig:colmap_test}
 \end{figure}

In Figure~\ref{fig:colmap_test}, we show the qualitative comparisons between two rendering options with PSNR on the corresponding images.
This reconstruction is sparse and fails to represent reflective surfaces and the sky.

\begin{table}[!htbp]
    \centering
    \scalebox{0.85}{
    \begin{tabular}{c||c|c}
    \toprule 
    Method & PSNR* (\texttt{train}) $\uparrow$ & PSNR* (\texttt{test}) $\uparrow$  \\
    \midrule  \midrule
    COLMAP (point) & 13.019 & 11.933  \\
    COLMAP (surfel) & 13.291 & 12.343  \\
 \bottomrule
 \end{tabular}
 }
    \cuthalfcaptionup
    \caption{Quantitative results for COLMAP. We discard invisible pixels (e.g., sky pixels that COLMAP fails to reconstruct) when computing the PSNR.}
    \label{tab:colmap_comparisons}
    \cuthalfcaptionup
    \cuthalfcaptionup
\end{table}

\section{Examples from our Datasets}

\begin{figure*}
     \centering
     
     \includegraphics[width=0.12\linewidth]{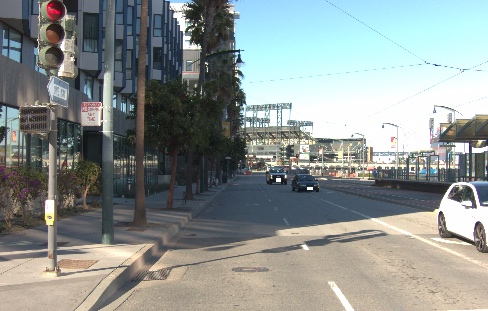}
     \includegraphics[width=0.12\linewidth]{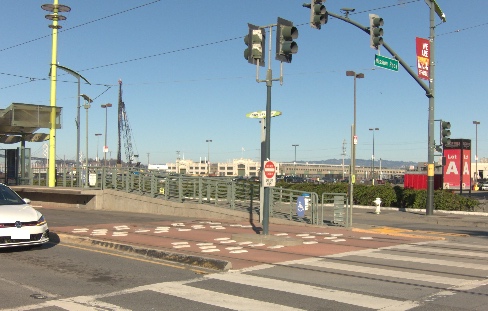}
     \includegraphics[width=0.12\linewidth]{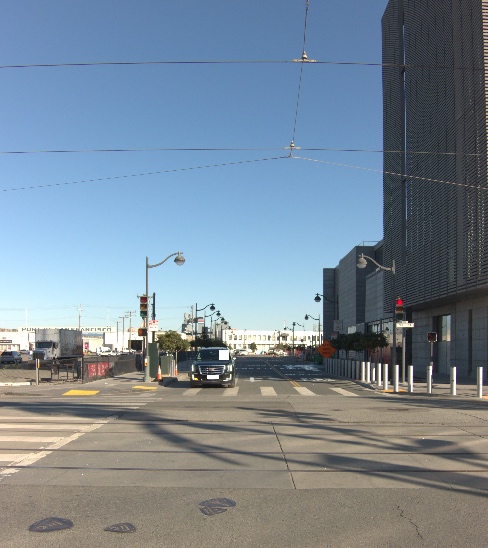}
     \includegraphics[width=0.12\linewidth]{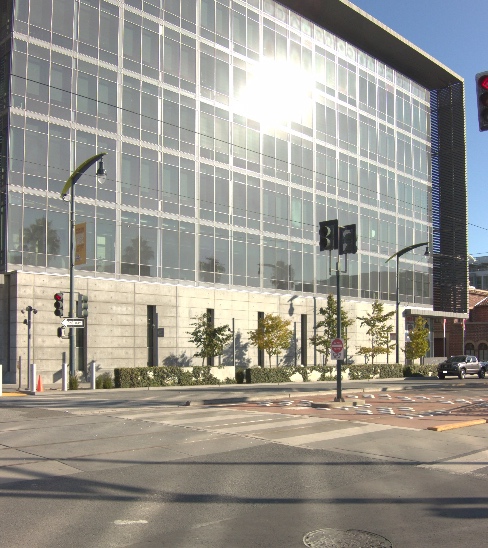}
     \includegraphics[width=0.12\linewidth]{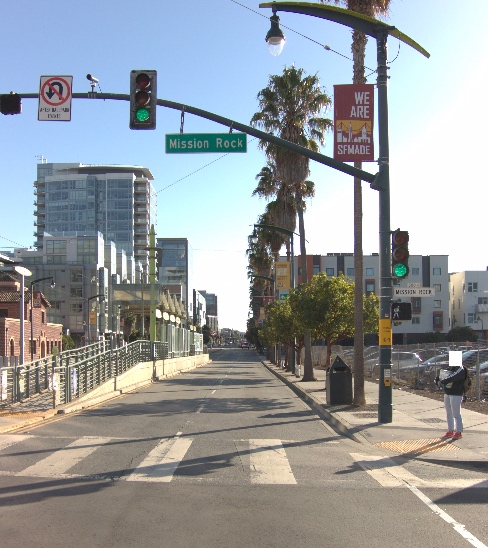}
     \includegraphics[width=0.12\linewidth]{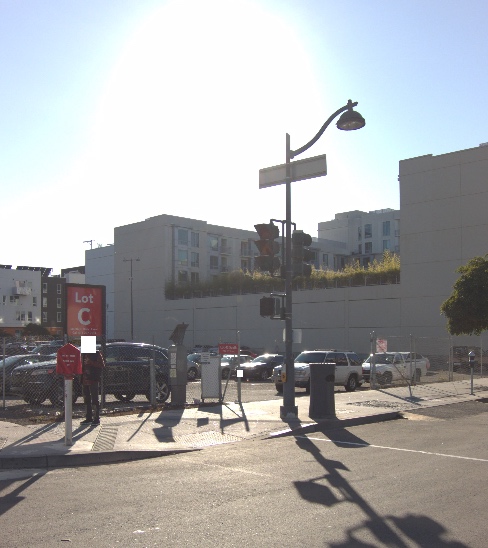}
     \includegraphics[width=0.12\linewidth]{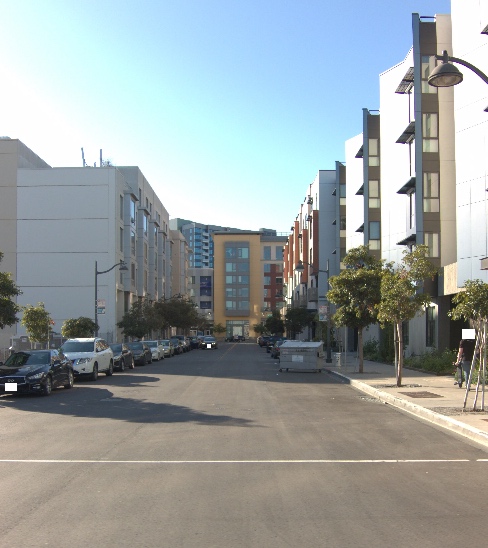}
     \includegraphics[width=0.12\linewidth]{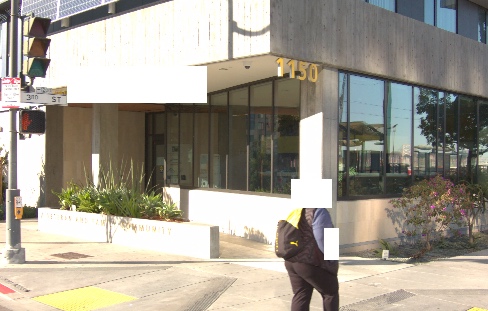}
     
     \includegraphics[width=0.12\linewidth]{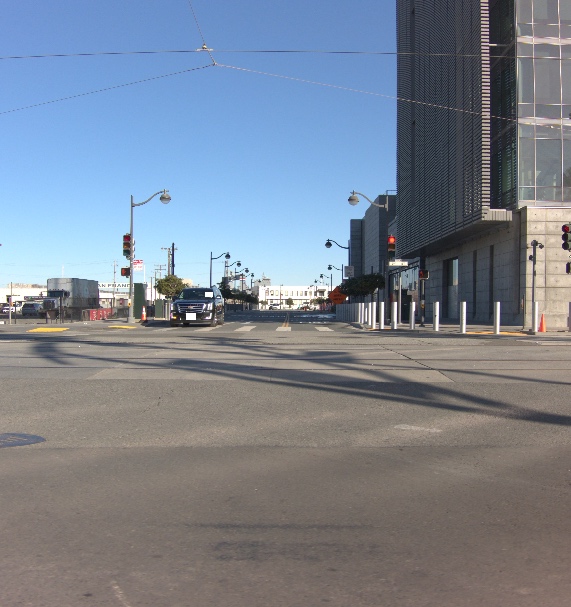}
     \includegraphics[width=0.12\linewidth]{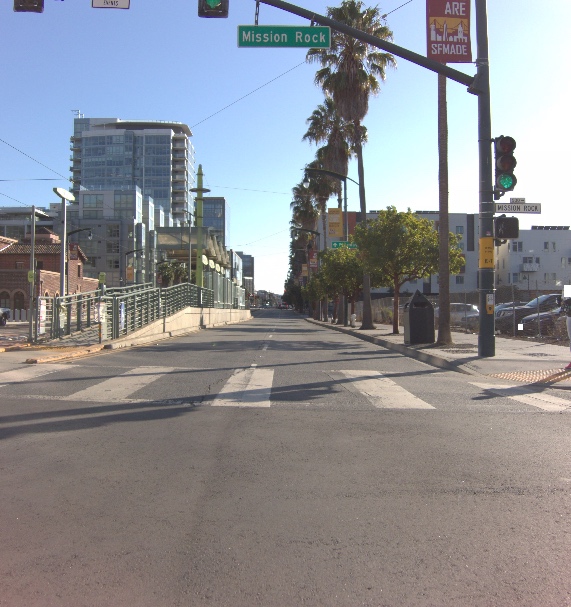}
     \includegraphics[width=0.12\linewidth]{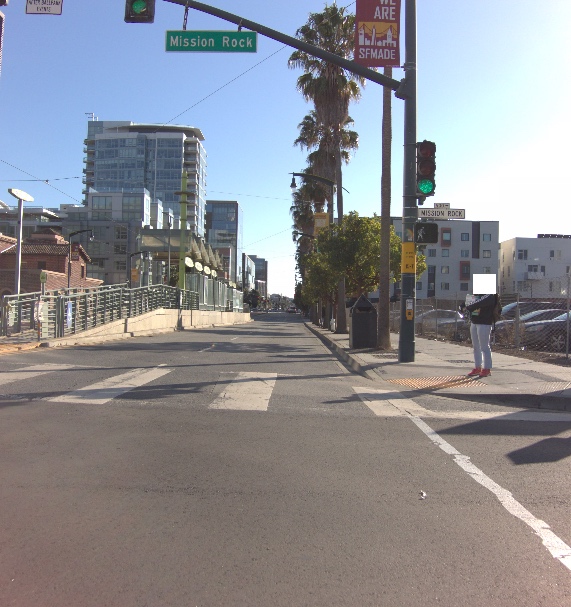}
     \includegraphics[width=0.12\linewidth]{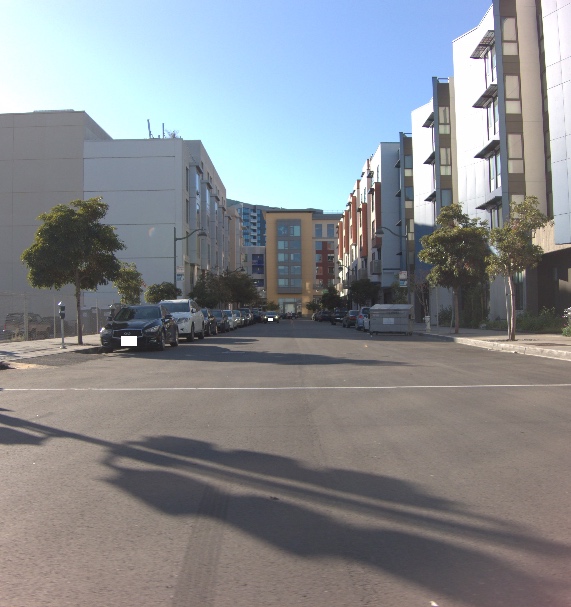}

     \cuthalfcaptionup
     \caption{Selection of images from our Mission Bay Dataset.}
     \cuthalfcaptionup
     \label{fig:mission_bay}
 \end{figure*}

\begin{figure*}
     \centering
     \includegraphics[width=1\linewidth]{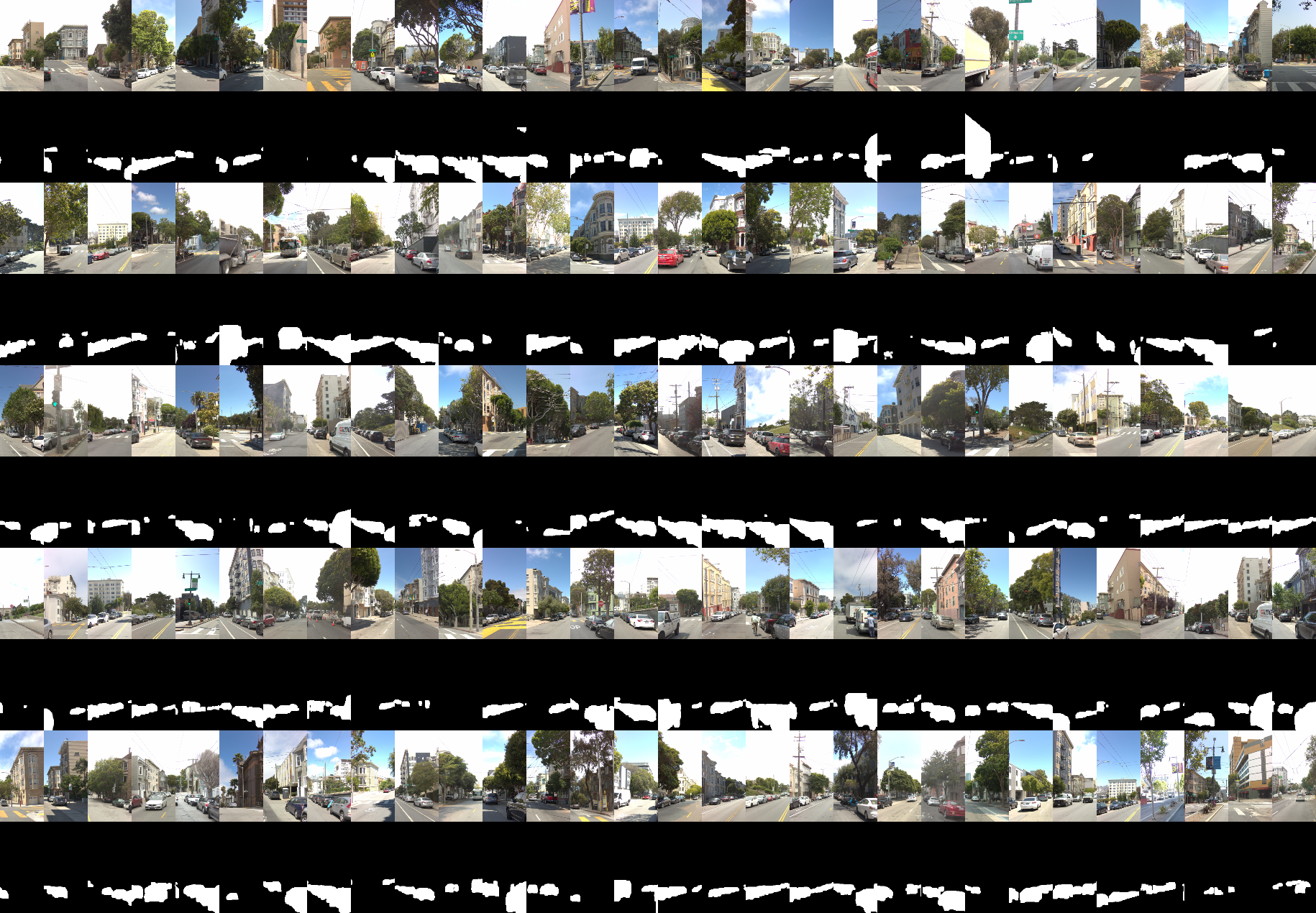}
     
     \cuthalfcaptionup
     \caption{Selection of front-facing images from our Alamo Square Dataset, alongside their transient object mask predicted by a pretrained semantic segmentation model.}
     \cuthalfcaptionup
     \label{fig:alamo_square}
 \end{figure*}

In Figure~\ref{fig:mission_bay}, we show the camera images from our Mission Bay dataset.
In Figure~\ref{fig:alamo_square}, we show both camera images and corresponding segmentation masks from our Alamo Square dataset.


\section{Societal Impact}

\subsection{Methodological}

Our method inherits the heavy compute footprint of NeRF models and we propose to apply them at an unprecedented scale. Our method also unlocks new use-cases for neural rendering, such as building detailed maps of the environment (mapping), which could cause more wide-spread use in favor of less computationally involved alternatives. Depending on the scale this work is being applied at, its compute demands can lead to or worsen environmental damage if the energy used for compute leads to increased carbon emissions. As mentioned in the paper, we foresee further work, such as caching methods, that could reduce the compute demands and thus mitigate the environmental damage.

\subsection{Application}

We apply our method to real city environments. During our own data collection efforts for this paper, we were careful to blur faces and sensitive information, such as license plates, and limited our driving to public roads. Future applications of this work might entail even larger data collection efforts, which raises further privacy concerns. While detailed imagery of public roads can already be found on services like Google Street View, our methodology could promote repeated and more regular scans of the environment. Several companies in the autonomous vehicle space are also known to perform regular area scans using their fleet of vehicles; however some might only utilize LiDAR scans which can be less sensitive than collecting camera imagery.
\end{appendices}

\end{document}